\documentclass{article} 
\usepackage{iclr2027_conference,times}

\iclrfinalcopy
\usepackage{amsmath,amsfonts,bm}

\def\eqref#1{equation~\ref{#1}}

\def\1{\bm{1}}

\DeclareMathAlphabet{\mathsfit}{\encodingdefault}{\sfdefault}{m}{sl}
\SetMathAlphabet{\mathsfit}{bold}{\encodingdefault}{\sfdefault}{bx}{n}

\usepackage{hyperref}
\usepackage{url}
\usepackage{booktabs}
\usepackage{amssymb}
\usepackage{wrapfig}
\usepackage{graphicx}
\usepackage{multirow}
\usepackage[table]{xcolor}
\usepackage{natbib}
\usepackage{array}
\newcolumntype{E}{>{\centering\arraybackslash}m{2.7em}}
\providecommand{\EvalBest}[1]{}
\providecommand{\EvalSecond}[1]{}
\providecommand{\EvalThird}[1]{}
\renewcommand{\EvalBest}[1]{\textbf{#1}}
\providecommand{\EvalRule}[2]{}
\renewcommand{\EvalRule}[2]{\begingroup
  \setbox0=\hbox{#2}%
  \dimen0=\dimexpr\dp0+0.45ex+0.4pt\relax
  \leavevmode\hbox{%
    \rlap{\lower\dimen0\hbox to \wd0{%
      \ifnum#1=0\relax
        \leaders\hrule height 0.4pt\hfill
      \else
        \cleaders\hbox{\kern0.12em\vrule width 0.38em height 0.4pt depth 0pt\kern0.12em}\hfill
      \fi
    }}%
    \box0}%
\endgroup}
\renewcommand{\EvalSecond}[1]{\EvalRule{0}{#1}}
\renewcommand{\EvalThird}[1]{\EvalRule{1}{#1}}

\providecommand{\EvalFitLabel}[4]{%
  \begingroup\setbox0=\hbox{#3}%
  \ifdim\wd0>#2\relax
    \expandafter\gdef\csname EvalShort#1\endcsname{1}#4%
  \else
    \expandafter\gdef\csname EvalShort#1\endcsname{0}#3%
  \fi\endgroup}
\providecommand{\EvalLabelNote}[3]{%
  \ifnum\csname EvalShort#1\endcsname=1\relax #2: #3;\space\fi}

\title{RED: Reconstruction Evolution Dynamics for Generalizable AI-Generated Image Detection}

\author{
Wenpeng Mu$^{1}$,
Junshan Jin$^{1}$,
Tanfeng Sun$^{1}$,
Xinghao Jiang$^{1}$,
and Qiang Xu$^{1\dagger}$\\
$^{1}$School of Computer Science, Shanghai Jiao Tong University, Shanghai, China\\
$^{\dagger}$Corresponding Author.\\
\texttt{wpmu@sjtu.edu.cn} \qquad \texttt{xuqiangwhu@sjtu.edu.cn}
}

\begin{document}

\maketitle

\begin{abstract}
The rapid evolution of image generators calls for forensic cues that generalize beyond known generation mechanisms. Existing detectors often rely on static image representations or endpoint reconstruction discrepancies, leaving the evolution of intermediate reconstruction stages underexplored. We observe that the relative token predictability of real and generated images can reverse across reconstruction scales, suggesting that intermediate stages may expose forensic evidence overlooked by endpoint comparisons. Motivated by this observation, we propose RED (Reconstruction Evolution Dynamics), a framework that captures transferable forensic cues from coarse-to-fine reconstruction evolution. To our knowledge, RED is the first framework to use scale-wise token predictability to guide forensic evidence aggregation across intermediate reconstruction states. It represents the reconstruction trajectory produced by a frozen multiscale VQ-VAE in the shared feature space of a frozen CLIP encoder. To connect the observed predictability variations with visual evidence, RED learns image-adaptive stage weights from scale-wise token negative log-likelihoods provided by a frozen VAR model. A cross-stage evidence aggregation module then jointly models the original-image representation and the weighted reconstruction features, capturing complementary forensic cues through interactions along the reconstruction trajectory. Experiments on six diverse benchmarks demonstrate that RED achieves the highest average accuracy of 92.5\% and average precision of 97.5\% among the evaluated methods. Further evaluations show strong robustness to common image degradations, supporting the value of reconstruction evolution for generalizable AI-generated image detection. \textit{The code will be made publicly available upon acceptance of this paper.}
\end{abstract}

\section{Introduction}

Advances in GANs~\cite{goodfellow2014gan}, diffusion models~\cite{ho2020ddpm}, and visual autoregressive models~\cite{tian2024var} have made realistic image synthesis widely accessible, bringing growing challenges to content authenticity and public trust. The continuous emergence of new generators and model versions requires detectors to remain reliable beyond the generation mechanisms represented in their training data. Generalizable AI-generated image detection therefore depends on identifying forensic cues that transfer across diverse and previously unseen generators.

Pretrained foundation models offer a strong starting point for this transfer. Detectors built on vision encoders and multimodal large language models (MLLMs) exploit representations learned from large-scale data to recognize generated images beyond known generators~\cite{ojha2023univfd, mu2025exda, li2026iapl,tan2026veritas}. However, separating generation artifacts from semantic correlations in the detection training data remains a challenge~\cite{guillaro2025biasfree,chen2025dda}. Reconstruction-based detection provides a complementary perspective by examining how an input image responds to a pretrained generative model. The resulting discrepancies provide an additional source of forensic evidence, but comparisons between the input and its final reconstruction emphasize the outcome of reconstruction, leaving the evolution of intermediate states underexplored.

The interpretation of these endpoint discrepancies varies across reconstruction settings. DIRE~\cite{wang2023dire} reports lower reconstruction errors for diffusion-generated images than for real images, whereas DID~\cite{qi2026did} observes more pronounced residuals for generated images in challenging settings. Error accumulation during reconstruction~\cite{luo2024lare2} and content- and frequency-dependent reconstruction difficulty~\cite{chu2025fire} further highlight the factors that shape these discrepancies. Together, these findings show that the magnitude of reconstruction error has no single, consistent interpretation across generators and reconstruction settings. An endpoint comparison summarizes the resulting difference, but does not describe how that difference develops as reconstruction progresses.

This leads to our central question: \textbf{What can the reconstruction process reveal that its endpoint cannot?} We investigate this question by analyzing real images and samples from 40 generators, analyzed with a pretrained visual autoregressive model (VAR) and its associated multiscale VQ-VAE. The VQ-VAE progressively incorporates quantized information across scales to construct a coarse-to-fine sequence of intermediate reconstructions. VAR evaluates the input-derived tokens through scale-wise negative log-likelihood (NLL), providing a measure of conditional token predictability at each stage of this hierarchy.

\begin{figure}[t]
    \centering

    \begin{minipage}[b]{0.48\linewidth}
        \centering
        \includegraphics[width=\linewidth]{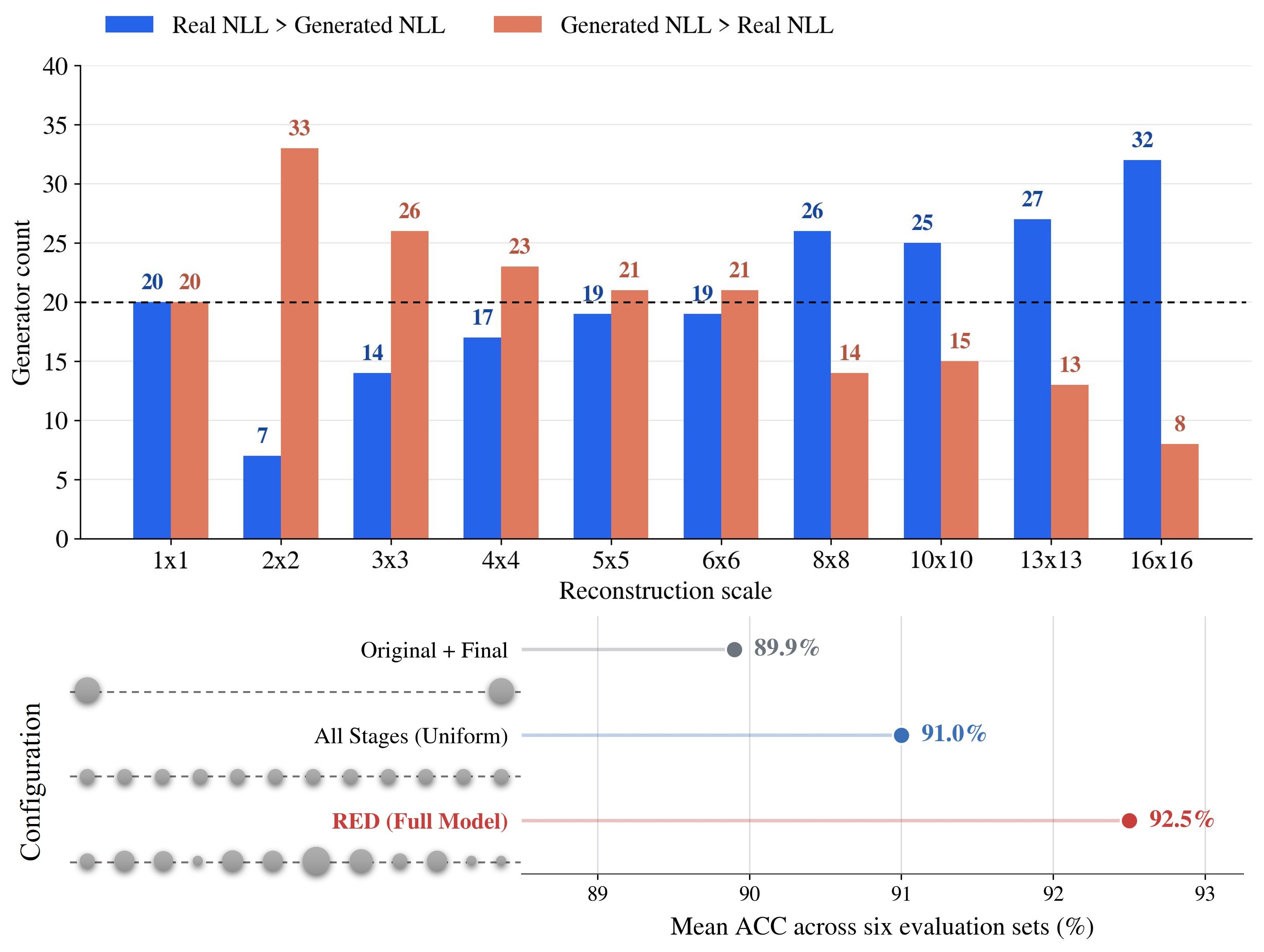}
        \caption{Scale-dependent token predictability and detection performance of RED.}
        \label{fig:motivation}
    \end{minipage}
    \hfill
    \begin{minipage}[b]{0.48\linewidth}
        \centering
        \includegraphics[width=\linewidth]{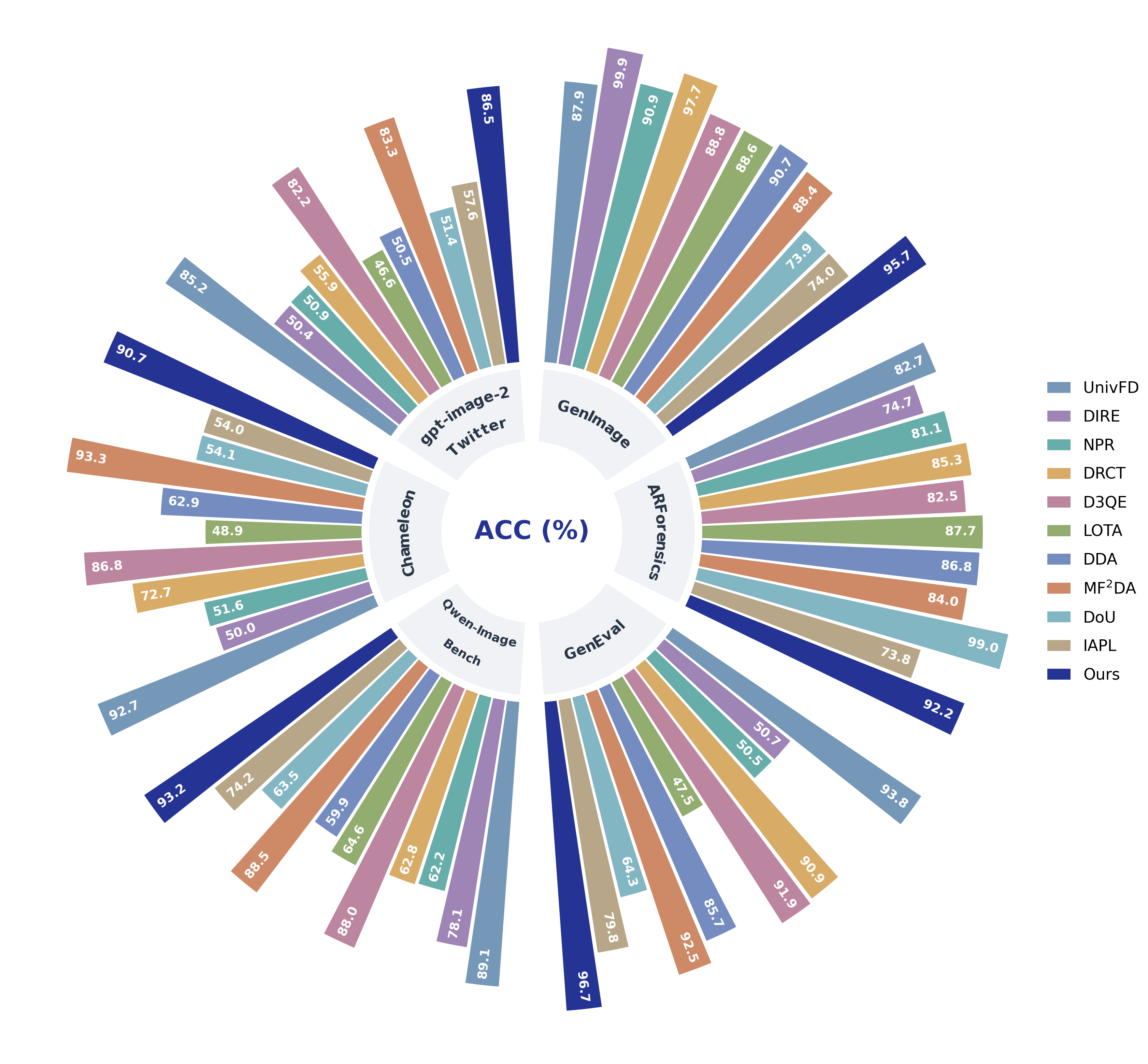}
        \caption{Detection accuracy comparison across six benchmarks (\%).}
        \label{fig:comparison}
    \end{minipage}

\end{figure}

As shown in the upper panel of Fig.~\ref{fig:motivation}, generated images have higher mean token NLL than real images for 33 of the 40 generators at the $2\times2$ scale, but for only 8 at the $16\times16$ scale. This reversal reveals a scale-dependent relationship between image authenticity and token predictability under the pretrained model that a final-scale score alone cannot describe. Further analysis of these scale-dependent NLL profiles are provided in Appendix~\ref{sec:nll_analysis}. The lower panel further shows that incorporating intermediate reconstruction states improves mean accuracy from 89.9\% for the original-plus-final baseline to 91.0\% with uniform weights, while adapting stage contributions through NLL-guided weighting in RED further improves accuracy to 92.5\%.

Building on this insight, we propose RED (\textbf{R}econstruction \textbf{E}volution \textbf{D}ynamics), a unified framework for capturing transferable forensic cues from coarse-to-fine reconstruction evolution. RED first establishes a \emph{reconstruction trajectory representation} by mapping the original image and successive reconstruction states into an ordered sequence in a shared visual feature space. \emph{NLL-guided stage weighting} translates scale-wise token predictability into image-specific stage contributions, learning how likelihood information should guide the allocation of visual evidence. \emph{Cross-stage evidence aggregation} then jointly contextualizes the original-image and weighted reconstruction representations to capture complementary evidence across the trajectory. The same stage weights guide both input feature modulation and output evidence pooling, coupling evidence allocation with its final readout. The resulting trajectory summary is combined with a dedicated, contextualized original-image readout for detection.

Experiments on GenImage~\cite{zhu2023genimage} and five out-of-domain benchmarks demonstrate RED's generalization across diverse generators and generation paradigms without target-domain adaptation, as shown in Fig.~\ref{fig:comparison}, RED achieves strong detection performance across all evaluated benchmarks, supporting its generalization across diverse generators and generation paradigms without target-domain adaptation.

Our contributions are summarized as follows:
\begin{itemize}
    \item \textbf{New perspective.}
    To our knowledge, we are the first to reveal a reversal in the dominant real--generated token-NLL ordering across reconstruction scales. Building on this finding, we introduce a forensic perspective that shifts attention from endpoint discrepancies to evidence evolving throughout coarse-to-fine reconstruction.

    \item \textbf{Unified framework.}
    We propose RED, which unifies NLL-guided adaptive stage weighting with cross-stage evidence aggregation. By using token predictability to adapt stage contributions and jointly modeling original-image and intermediate reconstruction representations, RED captures complementary forensic cues through image-specific weighting and cross-stage interactions.

    \item \textbf{Improved generalization.}
    Experiments on six benchmarks, including five out-of-domain benchmarks, demonstrate RED's generalization across diverse generators and generation paradigms without target-domain adaptation. RED achieves the highest average accuracy of 92.5\% and average precision of 97.5\% among the evaluated methods, with further evaluations demonstrating robustness to common image degradations.
\end{itemize}

\section{Related Work}

\textbf{AI-Generated Image Detection.}
Early work established the feasibility of cross-generator detection. CNNSpot~\cite{wang2020cnnspot} showed that suitable data augmentation enables a detector trained on a single CNN-based generator to generalize to other generators. Subsequent research expanded the representational basis of detection. UnivFD~\cite{ojha2023univfd} leveraged pretrained CLIP features to transfer visual knowledge from large-scale pretraining to image forensics. Beyond the choice of backbone, image-adaptive prompting~\cite{li2026iapl}, complementary feature fusion~\cite{mu2026mf2da}, and diversity-oriented representation learning~\cite{he2026dou} investigate how visual evidence can be adapted and combined for detection. MLLMs further extend this direction toward semantic anomaly analysis~\cite{tan2026anomagent} and interpretable authenticity reasoning~\cite{tan2026veritas}. Reconstruction provides a further means of probing an image through a pretrained generative model. Its role extends from supplying detection cues, as in the diffusion reconstruction errors used by DIRE~\cite{wang2023dire}, to constructing training examples, as in the reconstructed hard samples used for contrastive learning by DRCT~\cite{chen2024drct}.

\textbf{AI-Generated Image Detection Benchmarks.}
Generalization evaluation must account for differences in generation mechanisms, image content, and data sources~\cite{xu2025advancements}. GenImage provides a large-scale setting covering GANs and diffusion models across ImageNet~\cite{deng2009imagenet} categories. ARForensics~\cite{zhang2025d3qe} complements this coverage with visual autoregressive generators that employ different token prediction architectures, while EvalGEN~\cite{chen2025dda} includes more recent text-to-image generators spanning diffusion and autoregressive models. These benchmarks support evaluating transfer across diverse generator families and generation paradigms. Broader image distributions introduce additional challenges beyond architectural variation. Qwen-Image-Bench~\cite{Li2026QwenImageBenchFG}, originally developed for text-to-image evaluation, provides outputs from open-source and proprietary systems across varied creative scenarios. Chameleon~\cite{yan2025chameleon} includes high-quality real and synthetic images collected online, with generated samples curated to challenge human perception. GPT-Image-2-Twitter~\cite{Zewde2026GPTImage2IT} captures a complementary setting through user-shared images attributed to GPT-Image-2, spanning diverse subjects and visual styles.

\section{Method}

\begin{figure}
    \centering
    \includegraphics[width=\linewidth]{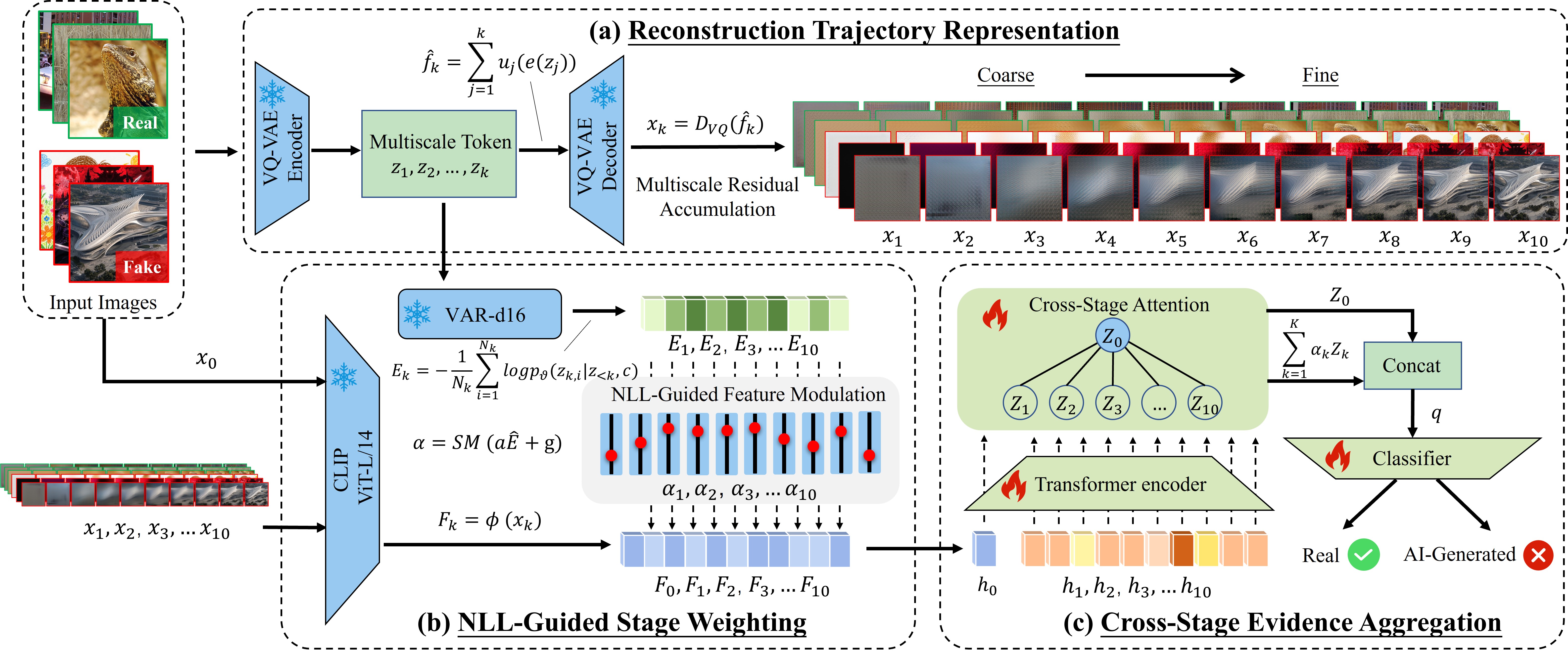}
    \caption{
    Overview of RED for generalizable AI-generated image detection.
    (a) Coarse-to-fine reconstruction states form a trajectory of visual evidence.
    (b) Scale-wise token NLLs guide image-specific weighting of reconstruction features.
    (c) Cross-stage evidence aggregation jointly models original-image and reconstruction representations, then combines a dedicated original-image readout with a trajectory summary pooled using the same stage weights.}
    \label{fig:framework}
\end{figure}

As illustrated in Fig.~\ref{fig:framework}, RED captures forensic evidence from coarse-to-fine reconstruction evolution through three components.

\textbf{How to Capture Evidence Throughout Reconstruction?}
RED constructs a reconstruction trajectory by progressively accumulating and decoding multiscale quantized representations with a frozen VQ-VAE. A frozen CLIP encoder maps the original image and successive reconstruction states into a shared visual feature space (Sec.~\ref{sec:dynamic}).

\textbf{How to Weight Stages for Each Image?}
RED maps scale-wise token NLLs from the frozen VAR model to image-specific stage weights, learning how token predictability should guide the contribution of visual evidence across reconstruction stages (Sec.~\ref{sec:nll}).

\textbf{How to Combine Evidence Across Stages?}
RED modulates the projected reconstruction features with the learned stage weights and jointly processes them with the unweighted original-image representation through cross-stage attention. Learnable stage embeddings retain token identity. The same weights are reused to pool the contextualized reconstruction features, and this summary is concatenated with the contextualized original-image representation for detection (Sec.~\ref{sec:detector}).

\subsection{Reconstruction Trajectory Representation}
\label{sec:dynamic}

We formulate reconstruction evolution as an ordered trajectory of cumulative reconstruction states, making intermediate visual evidence available for joint forensic analysis. Each state incorporates quantized information up to a particular scale, providing a structured representation of how image content develops throughout reconstruction. To construct this trajectory, we use the frozen multiscale VQ-VAE associated with VAR to extract a hierarchy of discrete token maps from an input image $x$:
\begin{equation}
(\mathbf{z}_1,\ldots,\mathbf{z}_K)
=
\mathcal{Q}_{\mathrm{ms}}
\bigl(\mathcal{E}_{\mathrm{VQ}}(x)\bigr),
\qquad
\mathbf{z}_k\in\{1,\ldots,V\}^{n_k\times n_k},
\label{eq:token_hierarchy}
\end{equation}
where $\mathcal{E}_{\mathrm{VQ}}$ is the image encoder, $\mathcal{Q}_{\mathrm{ms}}$ denotes multiscale residual quantization, and $V$ is the codebook size. We use $K=10$ scales, with $(n_1,\ldots,n_K)=(1,2,3,4,5,6,8,10,13,16)$.

We organize the quantized residual contributions into cumulative latent states and decode each state to obtain the reconstruction trajectory:
\begin{equation}
\begin{aligned}
\widehat{f}_k
&=
\sum_{j=1}^{k}
\mathcal{U}_j\bigl(e(\mathbf{z}_j)\bigr),
\qquad
x_k
=
\mathcal{D}_{\mathrm{VQ}}(\widehat{f}_k),\\
\mathcal{T}(x)
&=
(x_1,\ldots,x_K),
\qquad
k=1,\ldots,K,
\end{aligned}
\label{eq:reconstruction_trajectory}
\end{equation}
where $e$ denotes codebook lookup, $\mathcal{D}_{\mathrm{VQ}}$ is the shared decoder, and $\mathcal{U}_j$ resizes the token embeddings to the final latent resolution and applies the pretrained residual transformation. All states have the same output resolution, while successive states incorporate additional quantized residual information. The stage index therefore identifies the extent of accumulated reconstruction information, giving the trajectory a consistent coarse-to-fine organization.

A frozen CLIP ViT-L/14 encoder $\phi$ embeds the original image and all reconstruction states into a shared feature space:
\begin{equation}
F_k=\phi(x_k)\in\mathbb{R}^{d},
\qquad
\mathcal{F}(x)=(F_0,\ldots,F_K),
\qquad
x_0=x,
\label{eq:visual_trajectory}
\end{equation}
where $k=0,\ldots,K$. The ordered sequence $\mathcal{F}(x)$ retains the original-image representation alongside the visual evidence available at successive reconstruction stages. The same token hierarchy that constructs these states also supplies the likelihood evaluation in Sec.~\ref{sec:nll}, connecting visual trajectory representation and adaptive stage weighting through a common reconstruction process. Complete reconstruction sequences for real and generated images are visualized in Appendix~\ref{sec:reconstruction_visualization}.

\subsection{NLL-Guided Stage Weighting}
\label{sec:nll}

We translate the observed scale dependence of token predictability into a learnable rule for allocating reconstruction evidence. The central design is to use scale-wise NLLs to regulate the contributions of visual representations, with their influence learned through the detection objective.

Let $\mathbf{z}_{<k}=(\mathbf{z}_1,\ldots,\mathbf{z}_{k-1})$ denote the token maps from preceding scales. Under teacher forcing, the frozen VAR-d16 model evaluates the tokens at scale $k$ conditioned on these input-derived token maps and the unconditional conditioning input $c_{\varnothing}$. The token-averaged negative log-likelihood is
\begin{equation}
E_k(x)
=
-\frac{1}{N_k}
\sum_{i=1}^{N_k}
\log p_{\vartheta}
\bigl(z_{k,i}\mid\mathbf{z}_{<k},c_{\varnothing}\bigr),
\qquad
N_k=n_k^2,
\label{eq:stage_nll}
\end{equation}
where $\vartheta$ denotes the frozen VAR parameters and $z_{k,i}$ is the $i$-th token at scale $k$. Token averaging removes the direct dependence of the sum on token count. Larger $E_k$ indicates lower conditional token predictability under VAR.

To account for differences in the typical NLL ranges across scales, we standardize each score using the mean $\mu_k$ and standard deviation $\sigma_k$ estimated exclusively from the source training split. These statistics remain fixed during validation and testing. We then map the standardized scores to stage weights:
\begin{equation}
\begin{aligned}
\widehat{E}_k(x)
&=
\frac{E_k(x)-\mu_k}{\sigma_k},
\qquad
\widehat{\mathbf{E}}(x)
=
[\widehat{E}_1(x),\ldots,\widehat{E}_K(x)]^{\top},\\
\boldsymbol{\alpha}(x)
&=
\operatorname{softmax}
\bigl(a\widehat{\mathbf{E}}(x)+\mathbf{g}\bigr).
\end{aligned}
\label{eq:stage_weights}
\end{equation}

This parameterization combines image-dependent deviations in token predictability with stage preferences learned across the source training set. The shared scalar $a$ controls the direction and strength of the standardized NLL contribution, while the stage biases $\mathbf{g}\in\mathbb{R}^{K}$ capture preferences shared across images. The softmax normalizes these contributions over reconstruction stages, yielding $\alpha_k>0$ and $\sum_{k=1}^{K}\alpha_k=1$. Both $a$ and $\mathbf{g}$ are optimized for detection, allowing supervision to determine how the predictability signal should shape the allocation of visual evidence.

\subsection{Cross-Stage Evidence Aggregation}
\label{sec:detector}

We introduce cross-stage evidence aggregation to couple image-adaptive stage weighting with joint modeling of original-image and reconstruction representations. The aggregation design uses the same stage weights during both token construction and evidence readout, while retaining a dedicated original-image component in the final representation.

We first construct evidence tokens by modulating the projected reconstruction features with their stage weights and adding learnable stage embeddings:
\begin{equation}
h_k=
\begin{cases}
WF_0+\mathbf{p}_0, & k=0,\\
\alpha_k WF_k+\mathbf{p}_k, & 1\leq k\leq K,
\end{cases}
\label{eq:evidence_tokens}
\end{equation}
where $W\in\mathbb{R}^{d_h\times d}$ is a shared projection and $\mathbf{p}_k\in\mathbb{R}^{d_h}$ encodes token identity within the sequence, distinguishing the original image and individual reconstruction stages. Stage weights modulate the visual content entering cross-stage interaction, while the embeddings retain stage identity. The original-image token $h_0$ enters this interaction without NLL-based weighting.

Within the aggregation module, a Transformer encoder~\citep{vaswani2017transformer}, $\mathcal{A}_{\eta}$, jointly contextualizes the evidence tokens:
\begin{equation}
(Z_0,\ldots,Z_K)
=
\mathcal{A}_{\eta}(h_0,\ldots,h_K).
\label{eq:trajectory_aggregation}
\end{equation}
Self-attention enables each token to incorporate information from the original image and other reconstruction stages. Consequently, the resulting representations encode relationships across the trajectory in addition to the evidence contained in individual states.

For evidence readout, we reuse the stage weights to summarize the contextualized reconstruction tokens and concatenate this summary with the contextualized original-image token:
\begin{equation}
\begin{aligned}
q
&=
\operatorname{Concat}
\left(
Z_0,\,
\sum_{k=1}^{K}\alpha_k Z_k
\right),\\
\ell(x)
&=
f_{\omega}(q),
\qquad
s(x)=\sigma\bigl(\ell(x)\bigr),
\end{aligned}
\label{eq:detection}
\end{equation}
where $f_{\omega}$ is the classification head, $\ell(x)$ is its logit, and $\sigma$ is the sigmoid function. The score $s(x)\in(0,1)$ is used for ranking and thresholded detection, with $y=1$ denoting an AI-generated image.

Reusing $\boldsymbol{\alpha}$ connects the allocation of reconstruction evidence at the input to its contribution in the final readout: the same image-specific weights modulate the features presented to attention and pool its reconstruction outputs. The separate $Z_0$ component retains a dedicated original-image readout alongside the trajectory summary, with both informed by cross-stage interactions.

We jointly optimize stage weighting and evidence aggregation using binary cross-entropy with logits. For a minibatch $\mathcal{B}$, the objective is
\begin{equation}
\mathcal{L}
=
\frac{1}{|\mathcal{B}|}
\sum_{(x,y)\in\mathcal{B}}
\left[
\log\bigl(1+\exp(\ell(x))\bigr)-y\ell(x)
\right].
\label{eq:red_training}
\end{equation}
The VQ-VAE, VAR, and CLIP remain frozen. Training updates the stage-weighting parameters, shared projection, stage embeddings, Transformer encoder, and classification head, learning how reconstruction evidence is allocated, contextualized, and combined for detection.

\section{Experiments}
\label{sec:experiments}
\subsection{Experimental Settings}

\textbf{Baselines and Data.}
We compare RED with the 13 baselines in Table~\ref{tab:domain_dataset_summary}, all trained or fine-tuned on the same GenImage subset covering all training categories. Evaluation uses generated images from GenImage and five external collections listed in Table~\ref{tab:domain_dataset_summary}. All real images come from ImageNet, with matched real and generated sample counts within each evaluation split, an additional evaluation with alternative real-image sources is reported in Appendix~\ref{sec:additional_results} (Table~\ref{tab:real_source_generalization}). We report accuracy (ACC.) and average precision (A.P.), with generated images as the positive class.

\textbf{Model Configuration.}
RED uses frozen multiscale VQ-VAE, VAR-d16, and CLIP ViT-L/14 backbones, with $K=10$ reconstruction states at $256 \times 256$ resolution. The detection Transformer has two layers, a hidden dimension of $128$, four attention heads, and a feed-forward dimension of $256$. Transformer dropout and scale dropout are $0.1$ and $0.15$, respectively. VAR scores input-derived tokens under teacher forcing with unconditional conditioning $c=c_{\varnothing}$ and conditional dropout disabled. Token-averaged NLLs are standardized independently at each scale using source-training statistics, which remain fixed during validation and testing.

\textbf{Training and Evaluation.}
We optimize the stage-weighting and detection modules using binary cross-entropy with logits and AdamW, with learning rate and weight decay both set to $10^{-4}$, batch size $64$, and gradient clipping at a global norm of $1.0$. Training runs for at most $50$ epochs, with early-stopping patience of eight epochs. Checkpoints are selected by source-validation ACC at threshold $0.5$; the released checkpoint corresponds to epoch $16$. The same threshold is used for all RED evaluations, without target-domain training or calibration. Results are averaged equally across generators within each collection and across the six evaluation sets.

\subsection{Main Results}

Table~\ref{tab:domain_dataset_summary} summarizes performance across the six evaluation sets. RED achieves the highest average ACC of 92.5\% and AP of 97.5\%, exceeding the strongest baseline averages by 3.9 and 2.0 percentage points, respectively. Its ACC remains at least 86.5\% and AP at least 94.2\% across all sets, indicating consistently strong performance across the evaluated generated-image sources.

The comparison with DIRE and DRCT highlights the distinction between source-domain performance and generalization. On GenImage, these methods achieve 99.9\% and 97.7\% ACC, respectively, exceeding RED's 95.7\%. Across the five external evaluation sets, however, their average ACC falls to 60.8\% and 73.5\%, while RED reaches 91.9\%. Individual baselines retain advantages on specific sets, including Veritas and MF$^{2}$DA on Chameleon, but RED achieves the strongest overall results.

Table~\ref{tab:arforensics_geneval_acc} further demonstrates RED's balanced performance across generation mechanisms. RED ranks second on both ARForensics and EvalGEN, with ACC of 92.2\% and 96.7\%, respectively. In comparison, DoU leads on ARForensics with 99.0\% but obtains 64.3\% on EvalGEN, while Veritas achieves 99.6\% on EvalGEN and 77.2\% on ARForensics. RED also maintains ACC between 96.3\% and 96.9\% across all five EvalGEN generators, showing limited variation within this collection.

On Qwen-Image-Bench, RED achieves the highest average AP of 98.0\% across 18 generators, ranking first on 10 and maintaining AP between 96.0\% and 99.4\% (Table~\ref{tab:qwen_ap}). Veritas achieves higher ACC on this set (95.9\% versus 93.2\%), whereas RED achieves higher AP (98.0\% versus 95.6\%), distinguishing threshold-dependent classification performance from score-ranking performance. On GPT-Image-2-Twitter, RED achieves the best ACC of 86.5\% and AP of 94.2\%, exceeding UnivFD, the strongest baseline on this set, by 1.3 and 1.1 percentage points, respectively. Appendix~\ref{sec:additional_results} reports the remaining per-generator metrics (Tables~\ref{tab:genimage_results}, \ref{tab:arforensics_geneval_ap}, and~\ref{tab:qwen_acc}).

\begin{table}[t]
\centering
\caption{Comparison with state-of-the-art detectors on in-domain and out-of-domain benchmarks (\%). The best, second-best, and third-best results in each column are indicated by \EvalBest{bold}, \EvalSecond{solid underline}, and \EvalThird{dashed underline}, respectively. $\dagger$ denotes MLLM-based detectors.}
\label{tab:domain_dataset_summary}
\begingroup
\setlength{\tabcolsep}{2pt}
\renewcommand{\arraystretch}{1.15}
\resizebox{\linewidth}{!}{%
\begin{tabular}{@{}l|c|EE|*{5}{EE}|EE@{}}
\hline
\multirow{3}{*}{Method} & \multirow{3}{*}{Venue}
& \multicolumn{2}{c|}{In Domain} & \multicolumn{10}{c|}{Out of Domain} & \multicolumn{2}{c}{\multirow{2}{*}{Avg.}} \\
\cline{3-14}
& & \multicolumn{2}{c|}{GenImage} & \multicolumn{2}{c}{\EvalFitLabel{T0N0}{\dimexpr5.4em+2\tabcolsep\relax}{ARForensics}{ARF}}
& \multicolumn{2}{c}{\EvalFitLabel{T0N1}{\dimexpr5.4em+2\tabcolsep\relax}{EvalGEN}{EG}} & \multicolumn{2}{c}{\EvalFitLabel{T0N2}{\dimexpr5.4em+2\tabcolsep\relax}{Qwen-Image-Bench}{QIB}}
& \multicolumn{2}{c}{\EvalFitLabel{T0N3}{\dimexpr5.4em+2\tabcolsep\relax}{Chameleon}{CH}} & \multicolumn{2}{c|}{\EvalFitLabel{T0N4}{\dimexpr5.4em+2\tabcolsep\relax}{GPT-Image-2-Twitter}{G2-T}} & \multicolumn{2}{c}{} \\
\cline{3-16}
& & ACC. & A.P. & ACC. & A.P. & ACC. & A.P. & ACC. & A.P.
& ACC. & A.P. & ACC. & A.P. & ACC. & A.P. \\
\hline
CNNSpot~\cite{wang2020cnnspot} & CVPR & 61.2 & 66.9 & 61.9 & 66.3 & 75.8 & 84.8 & 54.8 & 58.1 & 56.8 & 61.3 & 55.4 & 57.5 & 61.0 & 65.8 \\
UnivFD~\cite{ojha2023univfd} & CVPR & 87.9 & 94.9 & 82.7 & 90.3 & \EvalThird{93.8} & \EvalThird{98.9} & \EvalThird{89.1} & 95.4 & \EvalThird{92.7} & \EvalSecond{98.2} & \EvalSecond{85.2} & \EvalSecond{93.1} & \EvalSecond{88.6} & 95.1 \\
DIRE~\cite{wang2023dire} & ICCV & \EvalBest{99.9} & \EvalBest{99.9} & 74.7 & 82.3 & 50.7 & 77.9 & 78.1 & 86.5 & 50.0 & 42.7 & 50.4 & 57.6 & 67.3 & 74.5 \\
NPR~\cite{tan2024npr} & CVPR & 90.9 & 97.6 & 81.1 & 91.8 & 50.5 & 42.5 & 62.2 & 70.6 & 51.6 & 43.3 & 50.9 & 47.9 & 64.5 & 65.6 \\
DRCT~\cite{chen2024drct} & ICML & \EvalSecond{97.7} & \EvalThird{98.5} & 85.3 & \EvalSecond{98.3} & 90.9 & 98.7 & 62.8 & 89.9 & 72.7 & 96.9 & 55.9 & 89.9 & 77.5 & \EvalThird{95.4} \\
D$^{3}$QE~\cite{zhang2025d3qe} & ICCV & 88.8 & 93.9 & 82.5 & 89.5 & 91.9 & 97.2 & 88.0 & 94.1 & 86.9 & 92.2 & 82.3 & 90.0 & 86.7 & 92.8 \\
LOTA~\cite{wang2025lota} & ICCV & 88.6 & 96.9 & \EvalThird{87.7} & 96.5 & 47.5 & 37.9 & 64.6 & 61.3 & 48.9 & 39.9 & 46.7 & 37.5 & 64.0 & 61.7 \\
DDA~\cite{chen2025dda} & NeurIPS & 90.7 & 97.4 & 86.8 & 95.8 & 85.7 & 95.7 & 59.9 & 75.0 & 62.9 & 82.8 & 50.5 & 56.1 & 72.7 & 83.8 \\
MF$^{2}$DA~\cite{mu2026mf2da} & TIFS & 88.4 & 95.4 & 84.0 & 92.2 & 92.5 & 98.1 & 88.5 & \EvalSecond{95.8} & \EvalSecond{93.3} & \EvalBest{99.2} & 83.3 & \EvalThird{92.1} & \EvalThird{88.3} & \EvalSecond{95.5} \\
DoU~\cite{he2026dou} & CVPR & 73.9 & 81.6 & \EvalBest{99.0} & \EvalBest{99.9} & 64.3 & 88.9 & 63.5 & 77.4 & 54.2 & 71.0 & 51.4 & 61.7 & 67.7 & 80.1 \\
IAPL~\cite{li2026iapl} & CVPR & 74.0 & 81.7 & 73.8 & 81.6 & 79.8 & 94.3 & 74.2 & 85.5 & 54.1 & 66.2 & 57.7 & 74.9 & 68.9 & 80.7 \\
\hline
AReason$^{\dagger}$~\cite{tan2026anomagent} & ICLR & 65.6 & 60.9 & 77.8 & 71.6 & 86.9 & 79.7 & 82.6 & 75.8 & 84.5 & 78.4 & 83.0 & 77.4 & 80.1 & 74.0 \\
Veritas$^{\dagger}$~\cite{tan2026veritas} & ICLR & 79.8 & 79.5 & 77.2 & 76.9 & \EvalBest{99.6} & \EvalBest{99.4} & \EvalBest{95.9} & \EvalThird{95.6} & \EvalBest{95.5} & 95.5 & \EvalThird{83.5} & 82.6 & \EvalSecond{88.6} & 88.2 \\
\hline
\textbf{Ours} & / & \EvalThird{95.7} & \EvalSecond{98.9} & \EvalSecond{92.2} & \EvalThird{97.4} & \EvalSecond{96.7} & \EvalSecond{99.2} & \EvalSecond{93.2} & \EvalBest{98.0} & 90.7 & \EvalThird{97.1} & \EvalBest{86.5} & \EvalBest{94.2} & \EvalBest{92.5} & \EvalBest{97.5} \\
\hline
\end{tabular}%
}
\endgroup
\par\smallskip
\begin{minipage}{\linewidth}
\footnotesize
\textit{Note.} \EvalLabelNote{T0N0}{ARF}{ARForensics}
\EvalLabelNote{T0N1}{EG}{EvalGEN}
\EvalLabelNote{T0N2}{QIB}{Qwen-Image-Bench}
\EvalLabelNote{T0N3}{CH}{Chameleon}
\EvalLabelNote{T0N4}{G2-T}{GPT-Image-2-Twitter}
\end{minipage}
\end{table}

\begin{table}[t]
\centering
\caption{Comparison of detection accuracy (ACC) on ARForensics and EvalGEN (\%).}
\label{tab:arforensics_geneval_acc}
\begingroup
\setlength{\tabcolsep}{2pt}
\renewcommand{\arraystretch}{1.15}
\resizebox{0.85\linewidth}{!}{%
\begin{tabular}{@{}l|*{7}{E}|E|*{5}{E}|E@{}}
\hline
\multirow{2}{*}{Method} & \multicolumn{8}{c|}{ARForensics (ACC.)} & \multicolumn{6}{c}{EvalGEN (ACC.)} \\
\cline{2-15}
& \EvalFitLabel{SplitACCN0}{2.7em}{Infinity}{Inf} & \EvalFitLabel{SplitACCN1}{2.7em}{Janus-Pro}{JP} & \EvalFitLabel{SplitACCN2}{2.7em}{LlamaGen}{LG} & \EvalFitLabel{SplitACCN3}{2.7em}{Open-MAGVIT2}{OMV2} & RAR & Switti & VAR & Avg. & Flux & GoT & \EvalFitLabel{SplitACCN10}{2.7em}{Infinity}{Inf} & NOVA & \EvalFitLabel{SplitACCN12}{2.7em}{OmniGen}{OG} & Avg. \\
\hline
CNNSpot & 75.9 & 74.3 & 59.0 & 49.6 & 54.0 & 72.2 & 48.4 & 61.9 & 72.4 & 79.9 & 67.4 & 79.1 & 80.2 & 75.8 \\
UnivFD & 94.7 & 94.1 & 73.5 & 67.1 & 91.9 & 92.0 & 65.7 & 82.7 & \EvalThird{92.6} & 94.0 & \EvalThird{94.1} & 94.5 & 93.7 & \EvalThird{93.8} \\
DIRE & 73.5 & 74.5 & 76.0 & 73.3 & 76.7 & 75.0 & 74.3 & 74.7 & 50.6 & 51.0 & 50.5 & 50.7 & 50.6 & 50.7 \\
NPR & 89.6 & 96.6 & 82.8 & 62.3 & 57.1 & 92.8 & \EvalThird{86.3} & 81.1 & 48.5 & 49.6 & 51.7 & 50.5 & 52.3 & 50.5 \\
DRCT & \EvalThird{98.0} & \EvalSecond{97.9} & \EvalThird{88.9} & 74.4 & 66.8 & 97.6 & 73.6 & 85.3 & 87.1 & \EvalSecond{97.3} & 86.1 & 87.7 & 96.1 & 90.9 \\
D$^{3}$QE & 93.2 & 92.4 & 74.4 & 68.0 & 90.7 & 90.2 & 68.8 & 82.5 & 90.7 & 92.2 & 92.1 & 92.1 & 92.2 & 91.9 \\
LOTA & 89.0 & 87.7 & 86.7 & \EvalThird{88.2} & 87.7 & 90.3 & 84.4 & \EvalThird{87.7} & 47.6 & 46.3 & 48.1 & 47.7 & 47.6 & 47.5 \\
DDA & 75.5 & \EvalThird{97.6} & \EvalSecond{95.2} & 71.0 & 82.0 & \EvalThird{98.2} & \EvalSecond{88.2} & 86.8 & 62.5 & 92.2 & 80.9 & \EvalThird{96.4} & \EvalThird{96.4} & 85.7 \\
MF$^{2}$DA & 93.0 & 93.4 & 72.7 & 72.7 & \EvalThird{92.0} & 91.3 & 72.9 & 84.0 & 91.4 & 94.0 & 92.9 & 92.5 & 91.9 & 92.5 \\
DoU & \EvalSecond{98.9} & \EvalBest{99.1} & \EvalBest{99.1} & \EvalBest{98.9} & \EvalBest{99.2} & \EvalBest{99.0} & \EvalBest{99.0} & \EvalBest{99.0} & 62.8 & 67.1 & 58.3 & 69.2 & 63.9 & 64.3 \\
IAPL & 74.1 & 74.3 & 73.7 & 72.4 & 74.6 & 73.5 & 73.9 & 73.8 & 70.7 & 78.0 & 81.0 & 90.4 & 78.9 & 79.8 \\
\hline
AReason$^{\dagger}$ & 89.5 & 81.0 & 80.0 & 70.0 & 77.5 & 80.5 & 66.0 & 77.8 & 86.5 & 83.5 & 89.5 & 86.5 & 88.5 & 86.9 \\
Veritas$^{\dagger}$ & \EvalBest{99.9} & 92.0 & 71.0 & 58.5 & 63.5 & \EvalSecond{98.5} & 57.0 & 77.2 & \EvalBest{99.9} & \EvalBest{99.5} & \EvalBest{99.0} & \EvalBest{99.5} & \EvalBest{99.9} & \EvalBest{99.6} \\
\hline
\textbf{Ours} & 96.7 & 96.3 & 85.6 & \EvalSecond{91.3} & \EvalSecond{96.3} & 96.3 & 82.7 & \EvalSecond{92.2} & \EvalSecond{96.3} & \EvalThird{96.7} & \EvalSecond{96.6} & \EvalSecond{96.8} & \EvalSecond{96.9} & \EvalSecond{96.7} \\
\hline
\end{tabular}%
}
\endgroup
\par\smallskip
\begin{minipage}{\linewidth}
\footnotesize
\textit{Note.} \EvalLabelNote{SplitACCN0}{Inf}{Infinity}
\EvalLabelNote{SplitACCN1}{JP}{Janus-Pro}
\EvalLabelNote{SplitACCN2}{LG}{LlamaGen}
\EvalLabelNote{SplitACCN3}{OMV2}{Open-MAGVIT2}
\EvalLabelNote{SplitACCN12}{OG}{OmniGen}
\end{minipage}
\end{table}

\begin{table}[t]
\centering
\caption{Comparison of average precision (AP) on Qwen-Image-Bench (\%).}
\label{tab:qwen_ap}
\begingroup
\setlength{\tabcolsep}{2pt}
\renewcommand{\arraystretch}{1.15}
\resizebox{\linewidth}{!}{%
\begin{tabular}{@{}l|*{18}{E}|E@{}}
\hline
\multirow{2}{*}{Method} & \multicolumn{19}{c}{Qwen-Image-Bench (A.P.)} \\
\cline{2-20}
& \EvalFitLabel{T4N0}{2.7em}{FLUX.2-pro}{F2-P} & \EvalFitLabel{T4N1}{2.7em}{FLUX.2\_max}{F2-M} & \EvalFitLabel{T4N2}{2.7em}{GLM-Image}{GLM} & \EvalFitLabel{T4N3}{2.7em}{GPT-Image-1}{G1} & \EvalFitLabel{T4N4}{2.7em}{GPT-Image-1.5}{G1.5} & \EvalFitLabel{T4N5}{2.7em}{HunyuanImage-3.0}{HY3} & \EvalFitLabel{T4N6}{2.7em}{Imagen-4.0}{I4} & \EvalFitLabel{T4N7}{2.7em}{Imagen-4.0-Ultra}{I4-U} & \EvalFitLabel{T4N8}{2.7em}{Qwen-Image}{QI} & \EvalFitLabel{T4N9}{2.7em}{Qwen-Image-2.0-pro}{Q2-P} & \EvalFitLabel{T4N10}{2.7em}{Qwen-Image-2512}{Q2512} & \EvalFitLabel{T4N11}{2.7em}{Seedream-4.0}{S4} & \EvalFitLabel{T4N12}{2.7em}{Seedream-4.5}{S4.5} & \EvalFitLabel{T4N13}{2.7em}{Seedream-5.0}{S5} & \EvalFitLabel{T4N14}{2.7em}{gpt-image-2}{G2} & \EvalFitLabel{T4N15}{2.7em}{kling\_v2\_1}{K2.1} & \EvalFitLabel{T4N16}{2.7em}{nano-banana-2.0}{NB2} & \EvalFitLabel{T4N17}{2.7em}{nano-banana-pro}{NB-P} & Avg. \\
\hline
CNNSpot & 57.7 & 51.7 & 67.0 & 62.1 & 58.4 & 55.4 & 63.0 & 61.6 & 68.9 & 53.1 & 60.6 & 56.4 & 56.6 & 52.9 & 55.9 & 62.4 & 51.4 & 51.5 & 58.1 \\
UnivFD & \EvalThird{96.0} & \EvalSecond{94.4} & 97.0 & 95.8 & 93.9 & 97.8 & 96.1 & 96.1 & 96.5 & 96.4 & 96.0 & \EvalThird{95.4} & \EvalThird{92.6} & \EvalThird{95.6} & 95.2 & 96.1 & 94.2 & 92.0 & 95.4 \\
DIRE & 90.7 & 64.6 & 81.0 & \EvalBest{99.9} & \EvalBest{99.9} & 92.6 & 95.3 & 95.0 & 82.1 & \EvalBest{99.9} & 66.0 & 75.9 & 77.0 & 85.7 & \EvalBest{99.9} & 79.2 & 72.2 & \EvalBest{99.9} & 86.5 \\
NPR & 53.8 & 59.3 & 45.2 & 95.1 & 83.3 & 80.8 & 82.3 & 83.0 & 69.8 & 93.7 & 43.8 & 59.2 & 70.4 & 42.5 & 86.1 & 59.6 & 74.1 & 89.3 & 70.6 \\
DRCT & 77.6 & 70.0 & \EvalBest{98.5} & 95.8 & 93.1 & \EvalSecond{98.5} & \EvalSecond{97.4} & \EvalBest{98.8} & 95.4 & 94.4 & 94.2 & 87.6 & 80.3 & 80.0 & 86.9 & 94.5 & 83.3 & 92.3 & 89.9 \\
D$^{3}$QE & 95.0 & 91.5 & 94.4 & 94.3 & 92.1 & 96.6 & 94.8 & 95.0 & 95.0 & 95.5 & 95.2 & 94.4 & 92.0 & 94.7 & 93.6 & 94.8 & 92.9 & 92.3 & 94.1 \\
LOTA & 37.6 & 48.7 & 41.3 & 96.2 & \EvalSecond{98.5} & 77.7 & 37.7 & 38.1 & 51.9 & \EvalThird{99.2} & 37.1 & 49.3 & 47.7 & 41.2 & \EvalSecond{98.6} & 50.4 & 53.2 & \EvalSecond{98.9} & 61.3 \\
DDA & 80.2 & 64.4 & 86.7 & 92.3 & 67.7 & 95.6 & 91.5 & 90.0 & 76.4 & 76.0 & 69.2 & 58.6 & 63.4 & 79.8 & 47.4 & 79.1 & 64.9 & 67.5 & 75.0 \\
MF$^{2}$DA & \EvalSecond{96.3} & \EvalThird{94.3} & 96.9 & 96.8 & 94.2 & \EvalThird{98.1} & 96.2 & \EvalThird{96.2} & \EvalThird{96.9} & 97.1 & \EvalThird{96.7} & \EvalSecond{95.9} & 91.7 & \EvalSecond{96.5} & 95.6 & \EvalThird{96.8} & \EvalSecond{95.6} & 92.8 & \EvalSecond{95.8} \\
DoU & 56.6 & 62.0 & 79.8 & 95.3 & \EvalThird{97.5} & 92.7 & 76.8 & 77.4 & 77.7 & 94.5 & 58.1 & 73.0 & 76.2 & 49.6 & 95.3 & 68.3 & 68.7 & 94.1 & 77.4 \\
IAPL & 59.7 & 74.4 & 80.4 & \EvalSecond{99.7} & \EvalBest{99.9} & 97.5 & 93.7 & 94.7 & 86.4 & \EvalSecond{99.4} & 60.1 & 83.4 & 82.6 & 66.5 & 97.7 & 82.4 & 81.6 & \EvalThird{98.5} & 85.5 \\
\hline
AReason$^{\dagger}$ & 79.1 & 74.0 & 83.6 & 78.2 & 64.2 & 81.5 & 83.8 & 77.8 & 82.2 & 77.0 & 74.5 & 67.3 & 72.7 & 77.4 & 70.0 & 86.6 & 68.2 & 67.0 & 75.8 \\
Veritas$^{\dagger}$ & 95.0 & 91.5 & \EvalThird{97.5} & \EvalThird{99.0} & 95.0 & 98.0 & \EvalThird{97.0} & 96.0 & \EvalSecond{98.0} & 96.0 & \EvalBest{99.0} & 92.5 & \EvalSecond{95.0} & 95.5 & 88.0 & \EvalSecond{98.0} & \EvalThird{95.5} & 94.0 & \EvalThird{95.6} \\
\hline
\textbf{Ours} & \EvalBest{97.8} & \EvalBest{96.0} & \EvalSecond{97.6} & 97.7 & 96.6 & \EvalBest{99.4} & \EvalBest{98.4} & \EvalSecond{98.5} & \EvalBest{99.2} & 98.9 & \EvalSecond{98.8} & \EvalBest{98.1} & \EvalBest{97.3} & \EvalBest{98.0} & \EvalThird{98.5} & \EvalBest{98.6} & \EvalBest{98.1} & 97.4 & \EvalBest{98.0} \\
\hline
\end{tabular}%
}
\endgroup
\par\smallskip
\begin{minipage}{\linewidth}
\footnotesize
\par\smallskip
\textit{Note.} \EvalLabelNote{T4N0}{F2-P}{FLUX.2-pro}
\EvalLabelNote{T4N1}{F2-M}{FLUX.2\_max}
\EvalLabelNote{T4N2}{GLM}{GLM-Image}
\EvalLabelNote{T4N3}{G1}{GPT-Image-1}
\EvalLabelNote{T4N4}{G1.5}{GPT-Image-1.5}
\EvalLabelNote{T4N5}{HY3}{HunyuanImage-3.0}
\EvalLabelNote{T4N6}{I4}{Imagen-4.0}
\EvalLabelNote{T4N7}{I4-U}{Imagen-4.0-Ultra}
\EvalLabelNote{T4N8}{QI}{Qwen-Image}
\EvalLabelNote{T4N9}{Q2-P}{Qwen-Image-2.0-pro}
\EvalLabelNote{T4N10}{Q2512}{Qwen-Image-2512}
\EvalLabelNote{T4N11}{S4}{Seedream-4.0}
\EvalLabelNote{T4N12}{S4.5}{Seedream-4.5}
\EvalLabelNote{T4N13}{S5}{Seedream-5.0}
\EvalLabelNote{T4N14}{G2}{gpt-image-2}
\EvalLabelNote{T4N15}{K2.1}{kling\_v2\_1}
\EvalLabelNote{T4N16}{NB2}{nano-banana-2.0}
\EvalLabelNote{T4N17}{NB-P}{nano-banana-pro}
\end{minipage}
\end{table}

\subsection{Robustness to Common Image Degradations}

We evaluate robustness on Qwen-Image-Bench, covering 18 generators. Four degradations are tested: JPEG compression with quality factor 50, Gaussian blur with $\sigma=1$, Gaussian noise with $\sigma=5$, and downsampling to $128\times128$ followed by upsampling to $256\times256$. All models are evaluated without additional fine-tuning, using mean average precision (mAP) across generators.

\begin{figure}
    \centering
    \includegraphics[width=\linewidth]{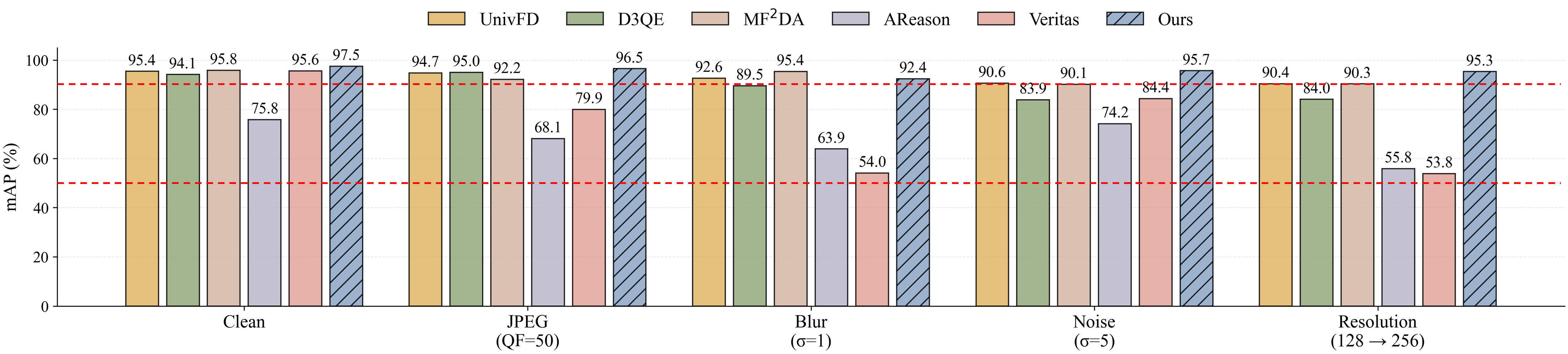}
    \caption{Robustness to common image degradations on Qwen-Image-Bench. Dashed lines mark the 90\% and 50\% mAP reference levels.}
    \label{fig:robust}
\end{figure}

As shown in Fig.~\ref{fig:robust}, RED achieves the highest mAP under JPEG compression, noise, and resolution reduction, reaching 96.5\%, 95.7\%, and 95.3\%, respectively. Averaged across all four degradations, RED achieves 95.0\% mAP, compared with 92.1\% for UnivFD and 92.0\% for MF$^{2}$DA. Blur remains the most challenging degradation for RED: its mAP decreases from 98.0\% on clean images to 92.4\%, below MF$^{2}$DA's 95.4\%. This sensitivity reflects the attenuation of fine-scale structures and textures by blurring, which can weaken the forensic evidence captured along the reconstruction trajectory.

The evaluated baselines exhibit markedly different sensitivities to image degradation. UnivFD and MF$^{2}$DA maintain at least 90.4\% and 90.1\% mAP, respectively, across all four conditions. In contrast, the evaluated MLLM-based detectors suffer substantial performance losses, revealing a pronounced gap between clean-image performance and degradation robustness. Most notably, Veritas achieves 95.6\% mAP on clean images but drops to 54.0\% under blur and 53.8\% under resolution reduction, corresponding to declines of 41.6 and 41.8 percentage points. AReason also decreases from 75.8\% to 55.8\% under resolution reduction, a loss of 20.0 points.

\subsection{Ablation Study}

Table~\ref{tab:ablation_study_optimized} evaluates the three components of RED: reconstruction trajectory representation, NLL-guided stage weighting, and cross-stage evidence aggregation. The ablations examine which reconstruction states provide useful evidence, how token NLLs should inform detection, and how evidence should be integrated across stages.

\textbf{Reconstruction Trajectory Representation.}
This ablation examines the value of intermediate reconstruction states beyond the original image and reconstruction endpoint. Original-image-only and final-reconstruction-only detection achieve average ACCs of 88.6\% and 83.0\%, respectively. The lower reconstruction-only accuracy may reflect the loss of useful forensic cues during reconstruction. Combining both representations improves ACC to 89.9\%, suggesting that reconstruction provides complementary evidence while the original image preserves cues lost in the process. Including all reconstruction stages with uniform weights further raises ACC to 91.0\%.

\textbf{NLL-Guided Stage Weighting.}
This ablation compares direct NLL features with NLL-guided stage weighting. NLL-only detection achieves 57.2\% average ACC, indicating limited standalone discrimination. Visual--NLL concatenation reaches 90.1\%, below uniform weighting at 91.0\%, suggesting that directly appending NLLs does not effectively exploit their information in this setting. To isolate image-dependent guidance from learned stage preferences, we fix $a=0$ while retaining learnable biases $\mathbf{g}$. The resulting static weights achieve 90.4\% average ACC, also below uniform weighting. Although static weighting outperforms RED on GenImage (97.8\% versus 95.7\%), RED performs better on all five out-of-domain sets and reaches 92.5\% overall, a gain of 2.1 percentage points. This pattern is consistent with fixed stage preferences being more suited to the source distribution and transferring less effectively.

\textbf{Cross-Stage Evidence Aggregation.}
This ablation compares the proposed aggregation design with pooling and MLP-based alternatives. Mean pooling achieves 81.1\% average ACC, while NLL-weighted averaging and MLP aggregation reach 91.2\% and 91.5\%, respectively. The weak mean-pooling result suggests that indiscriminate fusion may dilute useful stage-specific evidence. Full RED achieves 92.5\%, exceeding weighted averaging and MLP aggregation by 1.3 and 1.0 percentage points, respectively. These gains suggest additional value in jointly contextualizing stage representations within the complete aggregation design. Compared with weighted averaging, RED improves ACC by 3.4 points on Chameleon and 2.1 points on GPT-Image-2-Twitter, but decreases slightly by 0.2 points on ARForensics, indicating dataset-dependent benefits.

\begin{table}[t]
\centering
\caption{Component-wise ablation of RED across six evaluation sets (ACC, \%).}
\label{tab:ablation_study_optimized}
\begingroup
\setlength{\tabcolsep}{3pt}
\renewcommand{\arraystretch}{1.08}
\resizebox{\linewidth}{!}{%
\begin{tabular}{@{}llccccccc@{}}
\toprule
\multirow{2}{*}{Variant}
& \multirow{2}{*}{Modification / Setting}
& In Domain
& \multicolumn{5}{c}{Out of Domain}
& \multirow{2}{*}{Avg.} \\
\cmidrule(lr){3-3}
\cmidrule(lr){4-8}
& & GenImage & ARForensics & EvalGEN & QIB & Chameleon & G2-T & \\
\midrule

\rowcolor{gray!8}
\multicolumn{9}{@{}l}{
\textbf{\textit{Reconstruction Trajectory Representation}}
} \\
Original image only
& Remove all reconstruction inputs
& 87.9 & 82.7 & 93.8 & 89.1 & \textbf{92.7} & \underline{85.2} & 88.6 \\

Final reconstruction only
& Use the final reconstruction alone
& 88.0 & 79.1 & 93.2 & 79.5 & 85.8 & 72.7 & 83.0 \\

Original + final reconstruction
& Exclude intermediate reconstruction states
& 95.5 & 91.8 & 96.4 & 90.3 & 84.1 & 81.3 & 89.9 \\

All stages, uniform weights
& Use all reconstruction stages with equal weights
& 95.4 & 92.0 & 96.3 & 92.0 & 87.8 & 82.2 & 91.0 \\

\midrule
\rowcolor{gray!8}
\multicolumn{9}{@{}l}{
\textbf{\textit{NLL-Guided Stage Weighting}}
} \\
NLL-only detection
& Use NLL scores without visual features
& 60.9 & 56.5 & 68.4 & 52.5 & 53.0 & 51.6 & 57.2 \\

Visual--NLL concatenation
& Concatenate NLL scores with visual features
& 94.7 & 92.0 & 95.3 & 90.5 & 86.4 & 81.9 & 90.1 \\

Learned static weights
& Fix $a=0$; learn stage biases $\mathbf{g}$
& \textbf{97.8} & 91.3 & 94.9 & 88.1 & 87.6 & 82.9 & 90.4 \\

\midrule
\rowcolor{gray!8}
\multicolumn{9}{@{}l}{
\textbf{\textit{Cross-Stage Evidence Aggregation}}
} \\
Mean pooling
& Replace the aggregation design with mean pooling
& 86.6 & 80.5 & 93.5 & 75.7 & 82.5 & 67.8 & 81.1 \\

NLL-weighted averaging
& Replace the aggregation design with weighted averaging
& 95.3 & \textbf{92.4} & \underline{96.5} & 91.6 & 87.3 & 84.4 & 91.2 \\

MLP aggregation
& Replace the aggregation design with an MLP
& 95.6 & 92.0 & 96.2 & \underline{92.1} & 89.8 & 83.1 & \underline{91.5} \\

\midrule
\rowcolor{gray!8}
\multicolumn{9}{@{}l}{\textbf{\textit{Full Model}}} \\
RED
& Complete proposed configuration
& \underline{95.7} & \underline{92.2} & \textbf{96.7} & \textbf{93.2}
& \underline{90.7} & \textbf{86.5} & \textbf{92.5} \\
\bottomrule
\end{tabular}%
}

\smallskip
\begin{minipage}{\linewidth}
\footnotesize
QIB: Qwen-Image-Bench; G2-T: GPT-Image-2-Twitter.
\end{minipage}
\endgroup
\vspace{-2em}
\end{table}

\section{Conclusion}
We presented \textbf{RED} (\textbf{Reconstruction Evolution Dynamics}), a framework for generalizable AI-generated image detection that models visual representations across coarse-to-fine reconstruction stages. RED integrates multiscale VQ-VAE reconstruction, VAR-based NLL-guided stage weighting, and cross-stage evidence aggregation to combine original-image features with intermediate reconstruction evidence. Experiments on six diverse benchmarks show that RED achieves the highest average accuracy and average precision among the evaluated detectors, with strong generalization to unseen generators without target-domain adaptation. Further evaluations support its robustness to common image degradations. These findings highlight reconstruction evolution as a valuable source of complementary forensic evidence for generalizable AI-generated image detection.

\subsection*{AI Use Statement}

Generative AI tools were used to assist with experimental code development and language editing of the manuscript. The authors reviewed and verified all AI-assisted content and take full responsibility for the final work.

\bibliography{iclr2027_conference}
\bibliographystyle{iclr2027_conference}

\newpage
\appendix
\section{Dataset Construction and Reproducibility Details}
\label{sec:reproducibility}

\textbf{Source Data and Splits.}
Using sampling seed 2026, we select 1,000 images from each of eight generators in the GenImage training split: ADM, BigGAN, GLIDE, Midjourney, Stable Diffusion v1.4, Stable Diffusion v1.5, VQDM, and Wukong. Together with 8,000 ImageNet training images, these form a source dataset of 16,000 samples. A fixed 70/30 split, stratified across the eight generator groups and the real-image group, yields 11,200 training and 4,800 validation samples, with equal numbers of real and generated images in each split. Each generator contributes 700 training and 300 validation images. No ImageNet validation images are included in either source split. GenImage testing uses held-out samples from the same eight generators.

\textbf{Evaluation Data and Metric Aggregation.}
The main evaluation covers 40 generator/dataset groups across six collections, comprising 37,500 generated-image records (Table~\ref{tab:evaluation_sample_counts}). A fixed pool of 1,000 ImageNet validation images serves as the real-image reference. EvalGEN uses a fixed 500-image subset of this pool and 500 generated images per group; all other collections use the full reference pool and 1,000 generated images per group. The same real-image subset is reused across groups within each collection. We first macro-average metrics over groups within each collection and then compute the overall results as the unweighted mean of the six collection-level scores.

\begin{table}[htbp]
\centering
\caption{Per-group evaluation sample counts. Real images are drawn from a fixed pool of 1,000 ImageNet validation images; EvalGEN uses a fixed 500-image subset.}
\label{tab:evaluation_sample_counts}
\begin{tabular}{lrrr}
\toprule
Evaluation collection & Groups & Real/group & Generated/group \\
\midrule
GenImage & 8 & 1,000 & 1,000 \\
ARForensics & 7 & 1,000 & 1,000 \\
EvalGEN & 5 & 500 & 500 \\
Qwen-Image-Bench & 18 & 1,000 & 1,000 \\
Chameleon & 1 & 1,000 & 1,000 \\
GPT-Image-2-Twitter & 1 & 1,000 & 1,000 \\
\bottomrule
\end{tabular}
\end{table}

\textbf{Preprocessing and Feature Extraction.}
Following VAR's validation preprocessing, images are decoded as RGB, resized using Lanczos interpolation so that the shorter side is 288 pixels, and center-cropped to $256\times256$. They are then converted to tensors in $[0,1]$, without online random augmentation. The frozen multiscale VQ-VAE decodes ten cumulative reconstruction states at $256\times256$, using token-grid scales $(1,2,3,4,5,6,8,10,13,16)$. We use the same frozen CLIP ViT-L/14 encoder as UnivFD to map the original image and each reconstruction to a 768-dimensional feature vector. CLIP preprocessing applies a $224\times224$ center crop and channel-wise normalization with means $(0.48145466,0.45782750,0.40821073)$ and standard deviations $(0.26862954,0.26130258,0.27577711)$.

\textbf{NLL Computation and Model Selection.}
The official frozen VAR-d16 model evaluates input-derived tokens under teacher forcing, using the unconditional input $c=c_{\varnothing}$ with conditional dropout disabled. Tokens at the current scale serve only as prediction targets, while preceding scales provide the token conditioning. Token-averaged NLLs are standardized independently at each scale using statistics estimated exclusively from the source-training split and held fixed during validation and testing. RED contains 381,548 trainable parameters, while the VQ-VAE, VAR, and CLIP remain frozen. Training follows Section~\ref{sec:experiments}. Checkpoints are selected using source-validation ACC at a fixed decision threshold of $0.5$. No target-domain checkpoint selection, threshold calibration, or hyperparameter tuning is performed.

\textbf{Real-Image Source Control.}
To examine the effect of the real-image source, we pair a fixed set of 1,000 Chameleon generated images separately with 1,000 real images from ImageNet validation, Chameleon, LSUN, or MS COCO. The detector checkpoint, preprocessing, decision threshold, and NLL standardization statistics remain fixed, so that only the real-image source varies. Samples are selected without consulting detector scores or predictions. Real and generated images are not matched by content category because the sources do not share a common set of category annotations.

\section{Additional Experimental Results}
\label{sec:additional_results}

We complement the main evaluation with detailed per-generator results, reporting ACC and AP on GenImage, AP on ARForensics and EvalGEN, and ACC on Qwen-Image-Bench.

\textbf{Detailed Results on GenImage.}
Table~\ref{tab:genimage_results} shows that RED achieves an average ACC of 95.7\% and AP of 98.9\%. Although DIRE and DRCT obtain higher average ACC, RED maintains ACC between 93.2\% and 97.7\% across all eight generators. Its AP ranges from 98.0\% to 99.8\%, yielding the second-highest average AP on this benchmark.

\begin{table}[h]
\centering
\caption{Per-generator accuracy (ACC) and average precision (AP) on GenImage. All results are reported in percentages, and Avg. denotes the mean across eight generators.}
\label{tab:genimage_results}
\begingroup
\setlength{\tabcolsep}{2pt}
\renewcommand{\arraystretch}{1.15}
\resizebox{\linewidth}{!}{%
\begin{tabular}{@{}l|*{8}{EE}|EE@{}}
\hline
\multirow{3}{*}{Method} & \multicolumn{18}{c}{GenImage} \\
\cline{2-19}
& \multicolumn{2}{c}{ADM} & \multicolumn{2}{c}{\EvalFitLabel{T1N0}{\dimexpr5.4em+2\tabcolsep\relax}{BigGAN}{BG}}
& \multicolumn{2}{c}{Glide} & \multicolumn{2}{c}{\EvalFitLabel{T1N1}{\dimexpr5.4em+2\tabcolsep\relax}{Midjourney}{MJ}}
& \multicolumn{2}{c}{SD1.4} & \multicolumn{2}{c}{SD1.5}
& \multicolumn{2}{c}{VQDM} & \multicolumn{2}{c|}{\EvalFitLabel{T1N2}{\dimexpr5.4em+2\tabcolsep\relax}{Wukong}{WK}}
& \multicolumn{2}{c}{Avg.} \\
\cline{2-19}
& ACC. & A.P. & ACC. & A.P. & ACC. & A.P. & ACC. & A.P. & ACC. & A.P. & ACC. & A.P. & ACC. & A.P. & ACC. & A.P. & ACC. & A.P. \\
\hline
CNNSpot & 53.6 & 56.1 & 62.4 & 63.8 & 63.6 & 69.0 & 59.0 & 64.7 & 63.1 & 71.9 & 65.1 & 73.8 & 58.7 & 62.8 & 63.9 & 73.1 & 61.2 & 66.9 \\
UnivFD & 75.6 & 86.3 & 90.9 & 97.3 & 91.2 & 97.2 & 83.4 & 92.4 & 91.1 & 96.9 & 92.1 & 97.2 & 89.9 & 96.3 & 89.2 & 95.7 & 87.9 & 94.9 \\
DIRE & \EvalBest{99.9} & \EvalBest{99.9} & \EvalBest{99.9} & \EvalBest{99.9} & \EvalBest{99.9} & \EvalBest{99.9} & \EvalBest{99.9} & \EvalBest{99.9} & \EvalBest{99.9} & \EvalBest{99.9} & \EvalBest{99.9} & \EvalBest{99.9} & \EvalBest{99.8} & \EvalBest{99.9} & \EvalBest{99.9} & \EvalBest{99.9} & \EvalBest{99.9} & \EvalBest{99.9} \\
NPR & \EvalThird{93.2} & \EvalSecond{98.6} & 91.4 & 97.7 & 96.9 & \EvalThird{99.5} & 80.2 & 92.9 & 92.0 & \EvalThird{98.1} & 92.0 & 98.1 & 92.1 & 98.3 & 89.4 & 97.8 & 90.9 & 97.6 \\
DRCT & 92.1 & \EvalThird{97.7} & \EvalSecond{99.2} & \EvalThird{98.7} & \EvalSecond{97.9} & 98.8 & \EvalSecond{97.6} & \EvalThird{98.7} & \EvalSecond{99.0} & \EvalThird{98.1} & \EvalSecond{99.4} & \EvalSecond{99.3} & \EvalSecond{98.0} & 98.3 & \EvalSecond{99.0} & \EvalSecond{98.6} & \EvalSecond{97.7} & \EvalThird{98.5} \\
D$^{3}$QE & 83.4 & 89.9 & 92.3 & 96.1 & 90.5 & 94.6 & 84.1 & 91.4 & 91.0 & 95.7 & 90.2 & 94.7 & 90.2 & 95.3 & 88.9 & 93.9 & 88.8 & 93.9 \\
LOTA & 86.3 & 96.0 & 87.5 & 96.8 & 93.3 & 99.0 & 93.0 & \EvalSecond{98.8} & 88.6 & 97.0 & 88.4 & 96.8 & 85.8 & 95.5 & 86.3 & 95.6 & 88.6 & 96.9 \\
DDA & 89.6 & 97.4 & 89.8 & 97.9 & 88.2 & 97.0 & 88.9 & 97.3 & \EvalThird{98.3} & \EvalBest{99.9} & \EvalThird{98.3} & \EvalBest{99.9} & 74.1 & 89.4 & \EvalThird{98.2} & \EvalBest{99.9} & 90.7 & 97.4 \\
MF$^{2}$DA & 81.7 & 91.4 & 92.7 & 97.9 & 90.9 & 96.4 & 86.0 & 94.4 & 89.2 & 95.8 & 88.6 & 95.5 & 92.7 & \EvalThird{98.6} & 85.4 & 93.2 & 88.4 & 95.4 \\
DoU & 75.2 & 81.6 & 74.2 & 81.2 & 73.9 & 82.0 & 72.7 & 81.6 & 74.5 & 82.9 & 72.8 & 80.4 & 74.7 & 82.2 & 73.1 & 81.1 & 73.9 & 81.6 \\
IAPL & 72.8 & 80.2 & 75.2 & 83.1 & 76.0 & 83.5 & 72.0 & 80.6 & 73.1 & 80.7 & 74.3 & 82.2 & 75.2 & 82.4 & 73.7 & 81.0 & 74.0 & 81.7 \\
\hline
AReason$^{*}$ & 60.5 & 56.6 & 61.5 & 57.3 & 68.5 & 64.2 & 66.0 & 61.5 & 67.5 & 62.7 & 65.0 & 59.9 & 68.5 & 63.2 & 67.0 & 61.9 & 65.6 & 60.9 \\
Veritas$^{*}$ & 63.5 & 63.0 & 67.5 & 67.0 & 71.0 & 70.5 & 88.5 & 88.0 & 94.5 & 94.5 & 97.5 & 97.5 & 63.0 & 63.0 & 92.5 & 92.5 & 79.8 & 79.5 \\
\hline
\textbf{Ours} & \EvalSecond{94.8} & \EvalSecond{98.6} & \EvalThird{97.1} & \EvalSecond{99.4} & \EvalThird{97.7} & \EvalSecond{99.8} & \EvalThird{93.2} & 98.0 & 96.6 & \EvalSecond{99.2} & 96.6 & \EvalThird{99.2} & \EvalThird{95.2} & \EvalSecond{98.7} & 95.0 & \EvalThird{98.5} & \EvalThird{95.7} & \EvalSecond{98.9} \\
\hline
\end{tabular}%
}
\endgroup
\end{table}

\textbf{Average Precision on ARForensics and EvalGEN.}
As shown in Table~\ref{tab:arforensics_geneval_ap}, RED achieves 97.4\% average AP on ARForensics, ranking behind DoU and DRCT. On EvalGEN, it achieves the second-highest average AP of 99.2\%, following Veritas at 99.4\%. RED's AP remains between 98.9\% and 99.5\% across all five EvalGEN generators, demonstrating consistently strong performance on this benchmark.

\begin{table}[h]
\centering
\caption{Per-generator average precision (AP, \%) on ARForensics and EvalGEN. Avg. denotes the mean across generators within each benchmark.}
\label{tab:arforensics_geneval_ap}
\begingroup
\setlength{\tabcolsep}{2pt}
\renewcommand{\arraystretch}{1.15}
\resizebox{\linewidth}{!}{%
\begin{tabular}{@{}l|*{7}{E}|E|*{5}{E}|E@{}}
\hline
\multirow{2}{*}{Method} & \multicolumn{8}{c|}{ARForensics (A.P.)} & \multicolumn{6}{c}{EvalGEN (A.P.)} \\
\cline{2-15}
& \EvalFitLabel{SplitAPN0}{2.7em}{Infinity}{Inf.} & \EvalFitLabel{SplitAPN1}{2.7em}{Janus-Pro}{JP} & \EvalFitLabel{SplitAPN2}{2.7em}{LlamaGen}{LG} & \EvalFitLabel{SplitAPN3}{2.7em}{Open-MAGVIT2}{OMV2} & RAR & Switti & VAR & Avg. & Flux & GoT & \EvalFitLabel{SplitAPN10}{2.7em}{Infinity}{Inf.} & NOVA & \EvalFitLabel{SplitAPN12}{2.7em}{OmniGen}{OG} & Avg. \\
\hline
CNNSpot & 84.9 & 82.6 & 62.5 & 50.7 & 55.4 & 80.2 & 47.8 & 66.3 & 80.9 & 90.3 & 74.8 & 87.2 & 90.5 & 84.8 \\
UnivFD & 98.9 & 99.0 & 84.5 & 79.1 & 97.4 & 97.6 & 75.8 & 90.3 & \EvalThird{98.2} & \EvalBest{99.6} & \EvalBest{99.1} & 98.8 & 98.8 & \EvalThird{98.9} \\
DIRE & 81.2 & 81.7 & 83.1 & 81.3 & 84.0 & 82.9 & 81.8 & 82.3 & 77.9 & 80.4 & 71.1 & 80.7 & 79.5 & 77.9 \\
NPR & 97.7 & 99.5 & 95.1 & 83.1 & 72.0 & 98.5 & \EvalThird{96.4} & 91.8 & 35.5 & 35.2 & 49.0 & 44.7 & 48.2 & 42.5 \\
DRCT & \EvalThird{99.5} & 98.5 & \EvalThird{98.5} & \EvalSecond{97.8} & \EvalThird{98.0} & \EvalThird{99.1} & \EvalSecond{97.0} & \EvalSecond{98.3} & 97.9 & 98.4 & 98.4 & \EvalBest{99.7} & 99.1 & 98.7 \\
D$^{3}$QE & 96.9 & 97.5 & 83.1 & 78.8 & 96.2 & 95.2 & 78.8 & 89.5 & 95.0 & 96.8 & 97.8 & 98.2 & 98.0 & 97.2 \\
LOTA & 96.6 & 96.7 & 96.2 & 96.8 & 96.6 & 97.7 & 95.2 & 96.5 & 37.6 & 36.5 & 38.9 & 38.3 & 38.4 & 37.9 \\
DDA & 92.9 & \EvalSecond{99.8} & \EvalSecond{99.1} & 86.5 & 95.2 & \EvalBest{99.9} & \EvalSecond{97.0} & 95.8 & 85.8 & 98.8 & 95.1 & \EvalSecond{99.5} & \EvalThird{99.4} & 95.7 \\
MF$^{2}$DA & 98.7 & 98.7 & 83.6 & 84.7 & 97.7 & 97.5 & 84.4 & 92.2 & 97.7 & \EvalSecond{99.4} & \EvalThird{98.5} & 97.3 & 97.7 & 98.1 \\
DoU & \EvalBest{99.9} & \EvalBest{99.9} & \EvalBest{99.8} & \EvalBest{99.9} & \EvalBest{99.9} & \EvalBest{99.9} & \EvalBest{99.9} & \EvalBest{99.9} & 85.3 & 92.4 & 83.2 & 94.4 & 89.2 & 88.9 \\
IAPL & 81.1 & 82.3 & 81.5 & 81.5 & 82.3 & 81.5 & 81.1 & 81.6 & 90.9 & 93.6 & 95.4 & 97.9 & 93.8 & 94.3 \\
\hline
AReason$^{*}$ & 82.6 & 74.4 & 73.8 & 64.1 & 71.5 & 73.9 & 60.7 & 71.6 & 78.9 & 75.2 & 84.3 & 78.9 & 81.3 & 79.7 \\
Veritas$^{*}$ & \EvalBest{99.9} & 91.0 & 70.5 & 58.5 & 62.6 & 98.5 & 57.0 & 76.9 & \EvalBest{99.9} & 99.0 & \EvalSecond{99.0} & 99.0 & \EvalBest{99.9} & \EvalBest{99.4} \\
\hline
\textbf{Ours} & \EvalSecond{99.8} & \EvalThird{99.6} & 93.9 & \EvalThird{97.1} & \EvalSecond{99.5} & \EvalSecond{99.4} & 92.6 & \EvalThird{97.4} & \EvalSecond{98.9} & \EvalThird{99.3} & \EvalBest{99.1} & \EvalThird{99.4} & \EvalSecond{99.5} & \EvalSecond{99.2} \\
\hline
\end{tabular}%
}
\endgroup
\par\smallskip
\begin{minipage}{\linewidth}
\footnotesize
\par
\textit{Note.} \EvalLabelNote{SplitAPN0}{Inf.}{Infinity}
\EvalLabelNote{SplitAPN1}{JP}{Janus-Pro}
\EvalLabelNote{SplitAPN2}{LG}{LlamaGen}
\EvalLabelNote{SplitAPN3}{OMV2}{Open-MAGVIT2}
\EvalLabelNote{SplitAPN12}{OG}{OmniGen}
\end{minipage}
\end{table}

\textbf{Accuracy on Qwen-Image-Bench.}
Table~\ref{tab:qwen_acc} reports performance at the fixed decision threshold. RED achieves the second-highest average ACC of 93.2\%, behind Veritas at 95.9\%. Its ACC ranges from 88.9\% to 95.8\% across 18 generators and exceeds 90\% on 17 of them. These results complement the AP evaluation in the main text, showing that RED also maintains strong detection performance at its fixed operating threshold.

\begin{table}[h]
\centering
\caption{Per-generator accuracy (ACC, \%) on Qwen-Image-Bench. Avg. denotes the mean across 18 generators.}
\label{tab:qwen_acc}
\begingroup
\setlength{\tabcolsep}{2pt}
\renewcommand{\arraystretch}{1.15}
\resizebox{\linewidth}{!}{%
\begin{tabular}{@{}l|*{18}{E}|E@{}}
\hline
\multirow{2}{*}{Method} & \multicolumn{19}{c}{Qwen-Image-Bench (ACC.)} \\
\cline{2-20}
& \EvalFitLabel{T3N0}{2.7em}{FLUX.2-pro}{F2-P} & \EvalFitLabel{T3N1}{2.7em}{FLUX.2\_max}{F2-M} & \EvalFitLabel{T3N2}{2.7em}{GLM-Image}{GLM} & \EvalFitLabel{T3N3}{2.7em}{GPT-Image-1}{G1} & \EvalFitLabel{T3N4}{2.7em}{GPT-Image-1.5}{G1.5} & \EvalFitLabel{T3N5}{2.7em}{HunyuanImage-3.0}{HY3} & \EvalFitLabel{T3N6}{2.7em}{Imagen-4.0}{I4} & \EvalFitLabel{T3N7}{2.7em}{Imagen-4.0-Ultra}{I4-U} & \EvalFitLabel{T3N8}{2.7em}{Qwen-Image}{QI} & \EvalFitLabel{T3N9}{2.7em}{Qwen-Image-2.0-pro}{Q2-P} & \EvalFitLabel{T3N10}{2.7em}{Qwen-Image-2512}{Q2512} & \EvalFitLabel{T3N11}{2.7em}{Seedream-4.0}{S4} & \EvalFitLabel{T3N12}{2.7em}{Seedream-4.5}{S4.5} & \EvalFitLabel{T3N13}{2.7em}{Seedream-5.0}{S5} & \EvalFitLabel{T3N14}{2.7em}{gpt-image-2}{G2} & \EvalFitLabel{T3N15}{2.7em}{kling\_v2\_1}{K2.1} & \EvalFitLabel{T3N16}{2.7em}{nano-banana-2.0}{NB2} & \EvalFitLabel{T3N17}{2.7em}{nano-banana-pro}{NB-P} & Avg. \\
\hline
CNNSpot & 54.9 & 50.5 & 61.9 & 58.4 & 54.9 & 53.2 & 58.6 & 57.1 & 62.1 & 51.3 & 56.5 & 53.6 & 53.5 & 51.1 & 53.5 & 57.3 & 48.9 & 49.6 & 54.8 \\
UnivFD & \EvalThird{89.4} & \EvalThird{87.6} & \EvalThird{91.8} & 89.2 & 86.4 & \EvalThird{92.9} & \EvalThird{90.7} & \EvalThird{90.4} & \EvalThird{91.2} & 90.9 & \EvalThird{90.2} & \EvalThird{89.4} & 84.5 & \EvalThird{90.1} & 89.0 & 90.0 & 87.3 & 83.9 & \EvalThird{89.1} \\
DIRE & 88.1 & 61.9 & 62.4 & \EvalBest{99.9} & \EvalBest{99.8} & 88.2 & 88.6 & 88.2 & 62.5 & \EvalBest{99.9} & 50.8 & 62.2 & 62.6 & 63.2 & \EvalBest{99.9} & 62.3 & 65.2 & \EvalBest{99.9} & 78.1 \\
NPR & 52.5 & 54.4 & 51.1 & 84.9 & 65.1 & 65.0 & 70.3 & 71.1 & 58.5 & 81.7 & 50.1 & 52.4 & 55.2 & 49.3 & 71.2 & 55.5 & 58.5 & 73.9 & 62.2 \\
DRCT & 51.6 & 51.0 & 84.7 & 68.5 & 56.2 & 90.9 & 81.3 & 83.1 & 62.0 & 60.4 & 57.6 & 53.4 & 51.9 & 52.4 & 55.1 & 62.3 & 52.6 & 56.1 & 62.8 \\
D$^{3}$QE & 88.6 & 85.1 & 88.6 & 89.5 & 85.9 & 92.3 & 88.3 & 88.8 & 89.5 & 90.3 & 89.3 & 88.3 & \EvalThird{84.8} & 87.3 & 87.8 & 89.4 & 86.4 & 84.7 & 88.0 \\
LOTA & 46.5 & 55.9 & 49.1 & 87.7 & 91.8 & 75.7 & 46.5 & 47.0 & 58.7 & 94.9 & 46.2 & 55.3 & 54.2 & 48.4 & \EvalThird{93.1} & 57.7 & 61.0 & 93.8 & 64.6 \\
DDA & 61.6 & 53.5 & 66.0 & 74.2 & 53.3 & 82.0 & 72.3 & 71.9 & 54.9 & 57.6 & 52.7 & 51.3 & 52.1 & 58.7 & 50.2 & 59.2 & 52.6 & 54.6 & 59.9 \\
MF$^{2}$DA & 88.9 & 86.4 & 90.0 & 90.2 & 86.4 & 91.9 & 88.6 & 89.0 & 90.1 & 90.1 & 89.7 & 88.5 & 82.7 & 89.5 & 88.2 & 89.7 & \EvalThird{88.2} & 84.4 & 88.5 \\
DoU & 50.7 & 55.5 & 56.8 & 78.3 & 85.0 & 86.9 & 57.3 & 56.8 & 61.4 & 73.8 & 50.3 & 57.1 & 56.1 & 49.5 & 79.9 & 61.1 & 54.0 & 73.1 & 63.5 \\
IAPL & 50.8 & 58.6 & 59.9 & \EvalThird{98.3} & \EvalSecond{99.0} & 91.3 & 84.9 & 85.4 & 66.7 & \EvalSecond{96.5} & 50.7 & 64.0 & 60.5 & 52.4 & 92.2 & 63.9 & 66.8 & \EvalThird{93.9} & 74.2 \\
\hline
AReason$^{*}$ & 86.5 & 81.5 & 90.0 & 85.5 & 71.0 & 88.5 & 90.0 & 85.5 & 89.0 & 84.0 & 82.0 & 74.0 & 79.5 & 84.5 & 76.0 & \EvalThird{91.5} & 75.0 & 73.5 & 82.6 \\
Veritas$^{*}$ & \EvalBest{95.0} & \EvalBest{92.0} & \EvalBest{98.0} & \EvalSecond{99.0} & \EvalThird{95.5} & \EvalBest{99.0} & \EvalBest{97.5} & \EvalBest{96.5} & \EvalBest{98.0} & \EvalSecond{96.5} & \EvalBest{99.0} & \EvalSecond{92.5} & \EvalBest{95.5} & \EvalBest{95.5} & 88.5 & \EvalBest{98.0} & \EvalBest{96.5} & \EvalSecond{94.5} & \EvalBest{95.9} \\
\hline
\textbf{Ours} & \EvalSecond{92.7} & \EvalSecond{88.9} & \EvalSecond{92.4} & 92.5 & 90.2 & \EvalSecond{95.8} & \EvalSecond{93.6} & \EvalSecond{94.0} & \EvalSecond{95.5} & \EvalThird{95.2} & \EvalSecond{94.7} & \EvalBest{93.2} & \EvalSecond{91.8} & \EvalSecond{92.5} & \EvalSecond{94.6} & \EvalSecond{94.0} & \EvalSecond{93.5} & 92.0 & \EvalSecond{93.2} \\
\hline
\end{tabular}%
}
\endgroup
\par\smallskip
\begin{minipage}{\linewidth}
\footnotesize
\par\smallskip
\textit{Note.} \EvalLabelNote{T3N0}{F2-P}{FLUX.2-pro}
\EvalLabelNote{T3N1}{F2-M}{FLUX.2\_max}
\EvalLabelNote{T3N2}{GLM}{GLM-Image}
\EvalLabelNote{T3N3}{G1}{GPT-Image-1}
\EvalLabelNote{T3N4}{G1.5}{GPT-Image-1.5}
\EvalLabelNote{T3N5}{HY3}{HunyuanImage-3.0}
\EvalLabelNote{T3N6}{I4}{Imagen-4.0}
\EvalLabelNote{T3N7}{I4-U}{Imagen-4.0-Ultra}
\EvalLabelNote{T3N8}{QI}{Qwen-Image}
\EvalLabelNote{T3N9}{Q2-P}{Qwen-Image-2.0-pro}
\EvalLabelNote{T3N10}{Q2512}{Qwen-Image-2512}
\EvalLabelNote{T3N11}{S4}{Seedream-4.0}
\EvalLabelNote{T3N12}{S4.5}{Seedream-4.5}
\EvalLabelNote{T3N13}{S5}{Seedream-5.0}
\EvalLabelNote{T3N14}{G2}{gpt-image-2}
\EvalLabelNote{T3N15}{K2.1}{kling\_v2\_1}
\EvalLabelNote{T3N16}{NB2}{nano-banana-2.0}
\EvalLabelNote{T3N17}{NB-P}{nano-banana-pro}
\end{minipage}
\end{table}

\textbf{Generalization across real-image sources.}
Table~\ref{tab:real_source_generalization} compares RED with three baselines across four real-image sources while keeping the Chameleon-generated images fixed. RED achieves the highest average ACC of 89.73\% and AP of 96.45\%, exceeding the strongest baseline, MF$^{2}$DA, by 4.18 and 3.40 percentage points, respectively. Its ACC ranges from 88.5\% to 90.7\% and AP from 95.5\% to 97.1\%, showing limited variation across sources. Replacing ImageNet-val real images with Chameleon real images reduces RED's ACC by 2.2 points, compared with drops of 24.0--32.5 points for the baselines. On the Chameleon real-image source, RED exceeds MF$^{2}$DA by 19.5 points in ACC and 10.1 points in AP. Although MF$^{2}$DA performs better on ImageNet-val and achieves higher ACC on LSUN, RED maintains stronger overall performance and greater consistency. These results support improved robustness to real-image source changes under this evaluation protocol.

\begin{table}[t]
\centering
\caption{Comparison across real-image sources with a fixed
generated-image source. ACC and AP are reported in percentages.}
\label{tab:real_source_generalization}
\small
\setlength{\tabcolsep}{4.5pt}
\renewcommand{\arraystretch}{1.12}
\begin{tabular}{l*{4}{cc}|cc}
\toprule
\textbf{Method}
& \multicolumn{2}{c}{\textbf{ImageNet-val}}
& \multicolumn{2}{c}{\textbf{Chameleon}}
& \multicolumn{2}{c}{\textbf{LSUN}}
& \multicolumn{2}{c}{\textbf{MS COCO}}
& \multicolumn{2}{c}{\textbf{Avg.}} \\
\cmidrule(lr){2-3}
\cmidrule(lr){4-5}
\cmidrule(lr){6-7}
\cmidrule(lr){8-9}
\cmidrule(lr){10-11}
& ACC. & A.P. & ACC. & A.P. & ACC. & A.P. & ACC. & A.P. & ACC. & A.P. \\
\midrule
UnivFD
& 92.7 & 98.2
& 60.2 & 75.3
& 88.2 & 92.2
& 82.7 & 88.6
& 80.95 & 88.58 \\

D$^{3}$QE
& 86.9 & 92.2
& 62.9 & 67.6
& 88.3 & 94.1
& 87.3 & 94.2
& 81.35 & 87.03 \\

MF$^{2}$DA
& 93.3 & 99.2
& 69.0 & 85.4
& 90.0 & 93.7
& 89.9 & 93.9
& 85.55 & 93.05 \\

\midrule
\textbf{Ours}
& 90.7 & 97.1
& 88.5 & 95.5
& 89.6 & 96.3
& 90.1 & 96.9
& \textbf{89.73} & \textbf{96.45} \\
\bottomrule
\end{tabular}
\end{table}

\section{Scale-Wise NLL Analysis}
\label{sec:nll_analysis}

\begin{figure}
    \centering
    \includegraphics[width=0.78\linewidth]{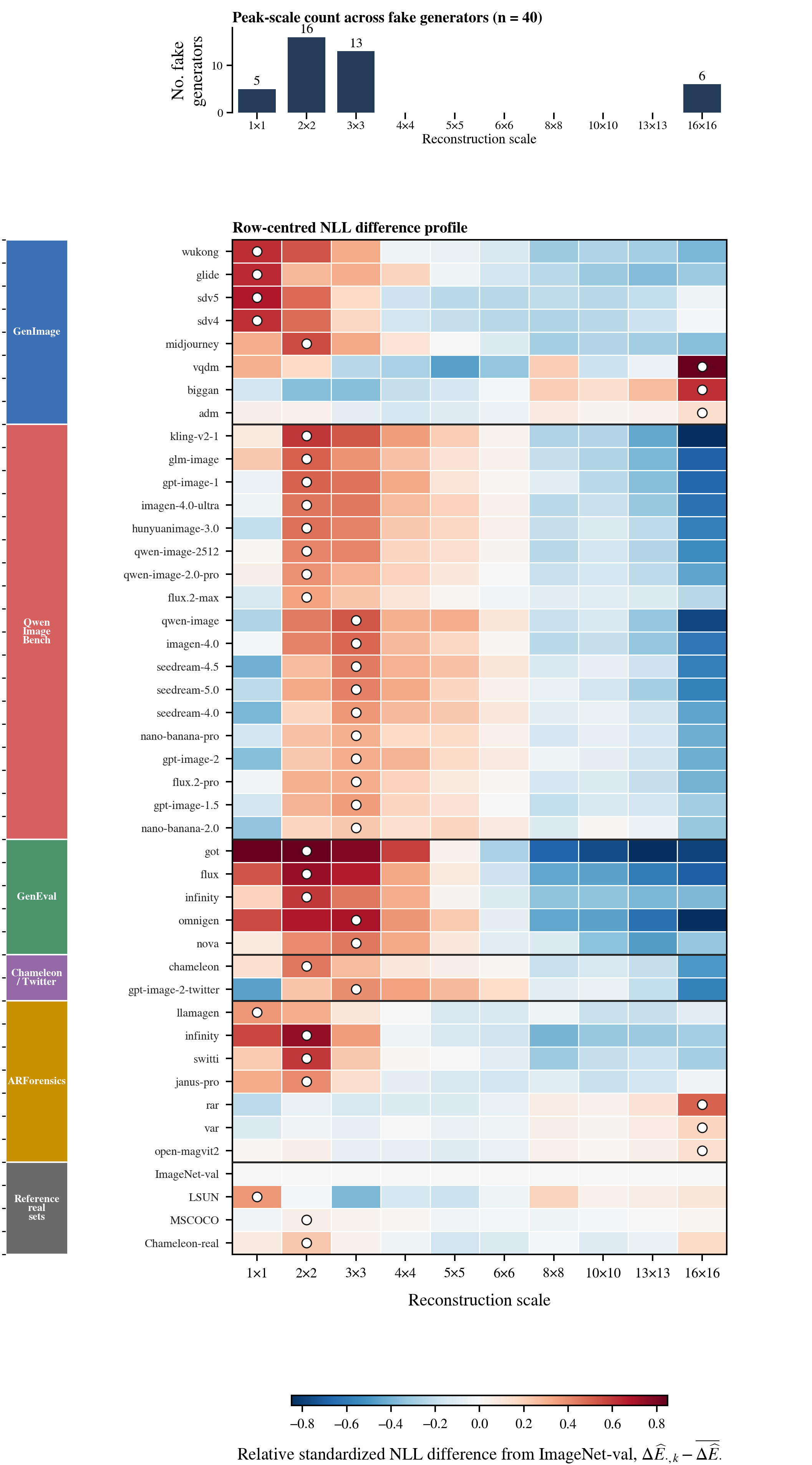}
    \caption{Scale-wise NLL difference profiles relative to ImageNet validation.}
    \label{fig:nll_heatmap}
\end{figure}

Fig.~\ref{fig:nll_heatmap} presents scale-wise profiles of standardized token-NLL differences relative to the ImageNet validation reference. For each image group, we compute the difference in mean standardized token NLL at each scale and subtract its average across scales. This row-centering highlights the shape of each profile: warm and cool colors indicate differences above and below the corresponding row mean, respectively. 

White circles mark the scale with the largest signed difference in each profile, and the upper panel summarizes these peak locations across the 40 generated-image groups. Five, sixteen, and thirteen groups peak at the $1\times1$, $2\times2$, and $3\times3$ scales, respectively, while the remaining six peak at $16\times16$. These final-scale peaks correspond to VQDM, BigGAN, ADM, RAR, VAR, and Open-MAGVIT2. The differing peak locations and profile shapes demonstrate that scale-wise NLL variation depends on the generator. These peaks characterize likelihood variation and do not directly identify the most discriminative reconstruction stages.

The additional real-image profiles illustrate variation across real-image sources, with the ImageNet validation profile equal to zero by construction. Together, these observations motivate examining likelihood information throughout reconstruction and learning how it should guide stage contributions, rather than relying on a single endpoint value or a fixed preferred scale.

\begin{figure}
    \centering
    \includegraphics[width=0.6\linewidth]{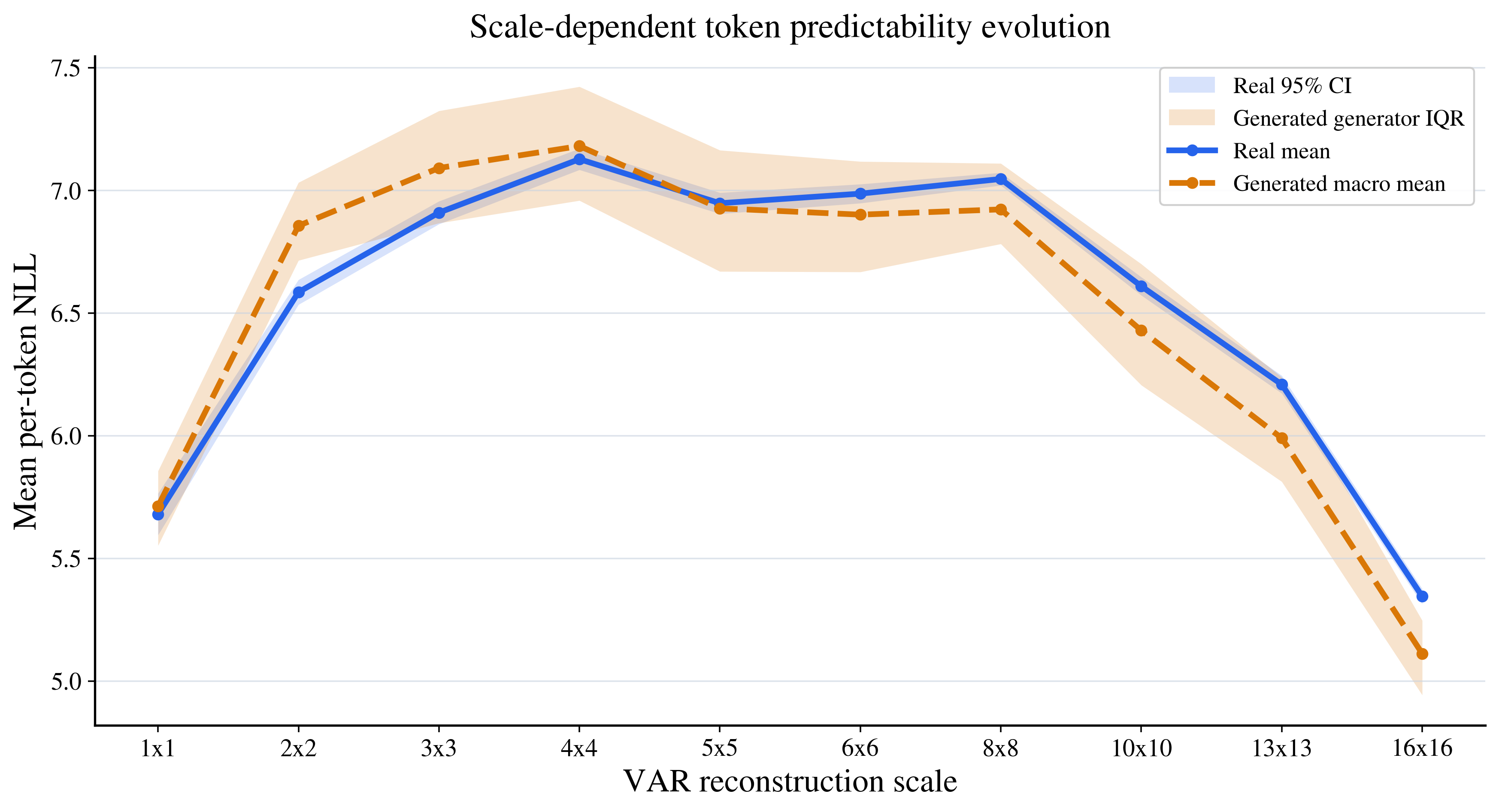}
    \caption{Scale-wise token-NLL profiles under the frozen VAR model.}
    \label{fig:nll_trajectory}
\end{figure}

\begin{figure}
    \centering
    \includegraphics[width=0.6\linewidth]{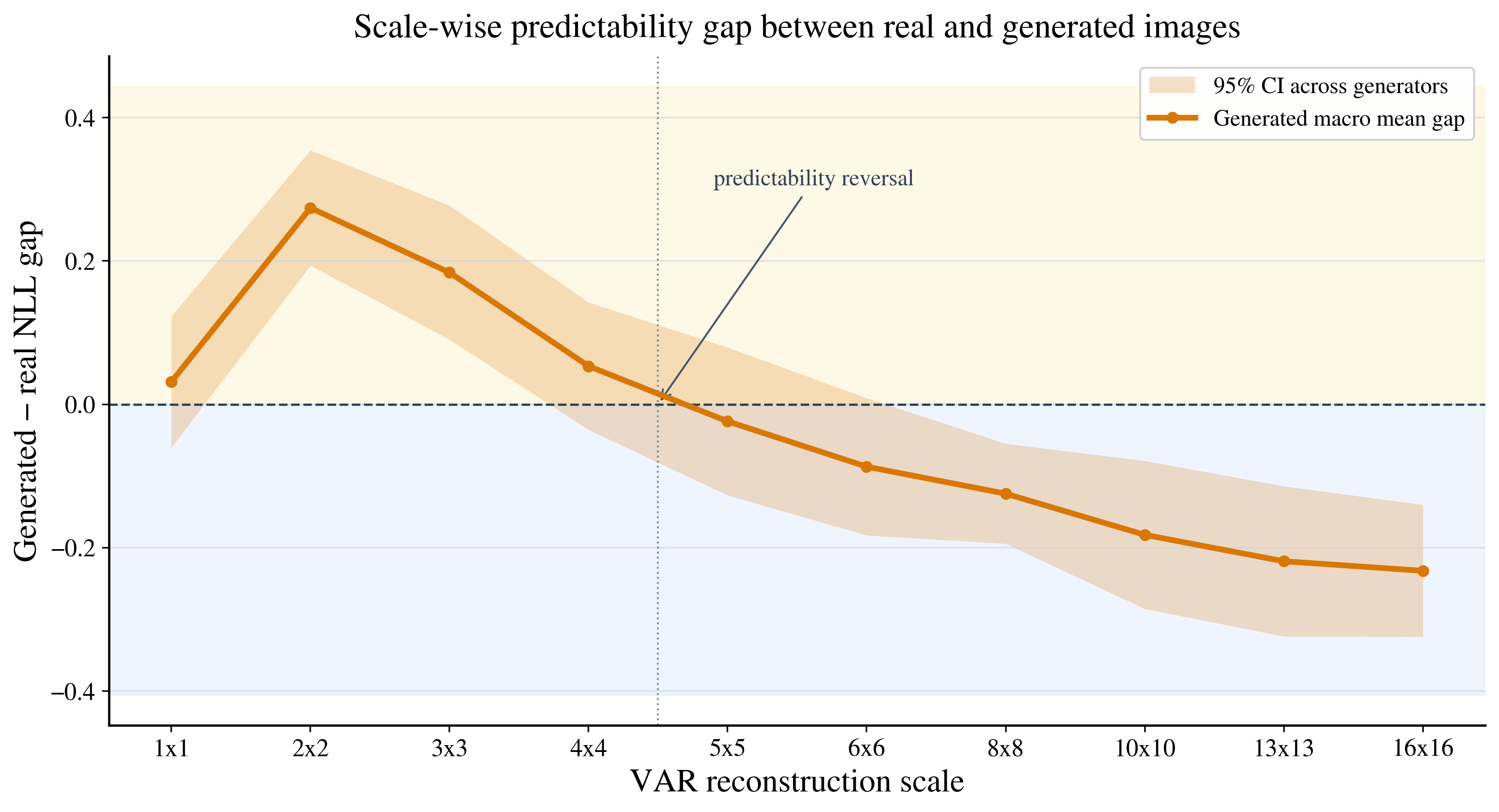}
    \caption{Scale-wise generated-minus-real token-NLL gaps.}
    \label{fig:nll_gap}
\end{figure}

Figures~\ref{fig:nll_trajectory} and~\ref{fig:nll_gap} examine token predictability across reconstruction scales under the frozen VAR model. We compare the real-image mean NLL with the macro-average of generator-group mean NLLs. The scale-wise gap is defined as
$\Delta_k=\bar{E}^{\mathrm{gen}}_k-\bar{E}^{\mathrm{real}}_k$:
positive values indicate lower conditional token predictability for generated images, while negative values indicate the reverse.

\textbf{Scale-dependent predictability reversal.}
The macro-mean gap is positive at coarse scales, particularly $2\times2$ and $3\times3$, changes sign between $4\times4$ and $5\times5$, and becomes negative at finer scales. This reversal shows that the relationship between authenticity and token predictability depends on the reconstruction stage. A final-scale score therefore omits the opposite ordering observed at earlier stages. The generated-image interquartile range also reveals variation across generator groups, indicating that the aggregate trend does not imply identical behavior for every generator.

\textbf{Connection to RED.}
The NLL profile and reconstruction trajectory arise from the same token hierarchy: $E_k$ measures the predictability of newly introduced tokens conditioned on preceding scales, while $x_k$ incorporates their contribution into the cumulative reconstruction. This correspondence provides stage-aligned predictability information for the visual representations. RED standardizes NLLs using source-training statistics and learns image-dependent weights through
$\boldsymbol{\alpha}(x)=\operatorname{softmax}(a\widehat{\mathbf{E}}(x)+\mathbf{g})$.
These weights modulate reconstruction features before cross-stage attention and pool the contextualized outputs, connecting evidence allocation with the final readout. The plotted population gaps motivate considering the full scale profile; they do not directly determine stage importance or prescribe a fixed weighting rule.

\textbf{Complementary evidence from ablations.}
Table~\ref{tab:ablation_study_optimized} connects these observations to detection performance. Adding intermediate states with uniform weights improves average ACC from 89.9\% to 91.0\%, supporting the value of intermediate visual evidence. Replacing learned static weights with image-dependent NLL guidance further improves ACC from 90.4\% to 92.5\%, with gains on all five out-of-domain sets. Together, these results support combining multistage visual evidence with adaptive predictability guidance, while the aggregate NLL profiles alone do not establish which stages are most discriminative for individual images.

\section{Qualitative Visualization of Reconstruction Trajectories}
\label{sec:reconstruction_visualization}

We visualize coarse-to-fine reconstruction trajectories for representative samples from all four real-image sources and six generated-image evaluation collections used in this work. For each sample, the frozen multiscale VQ-VAE used in RED produces ten successive reconstruction states by progressively accumulating and decoding quantized residual information. All states share a common output resolution, while successive stages incorporate additional information, progressing from coarse image structure toward finer visual detail. These examples illustrate reconstruction evolution across diverse image sources and provide an intuitive view of the intermediate states underlying RED's trajectory representation, complementing the quantitative scale-wise NLL analysis.

\begin{figure}[htbp]
    \centering
    \includegraphics[width=0.8\linewidth]{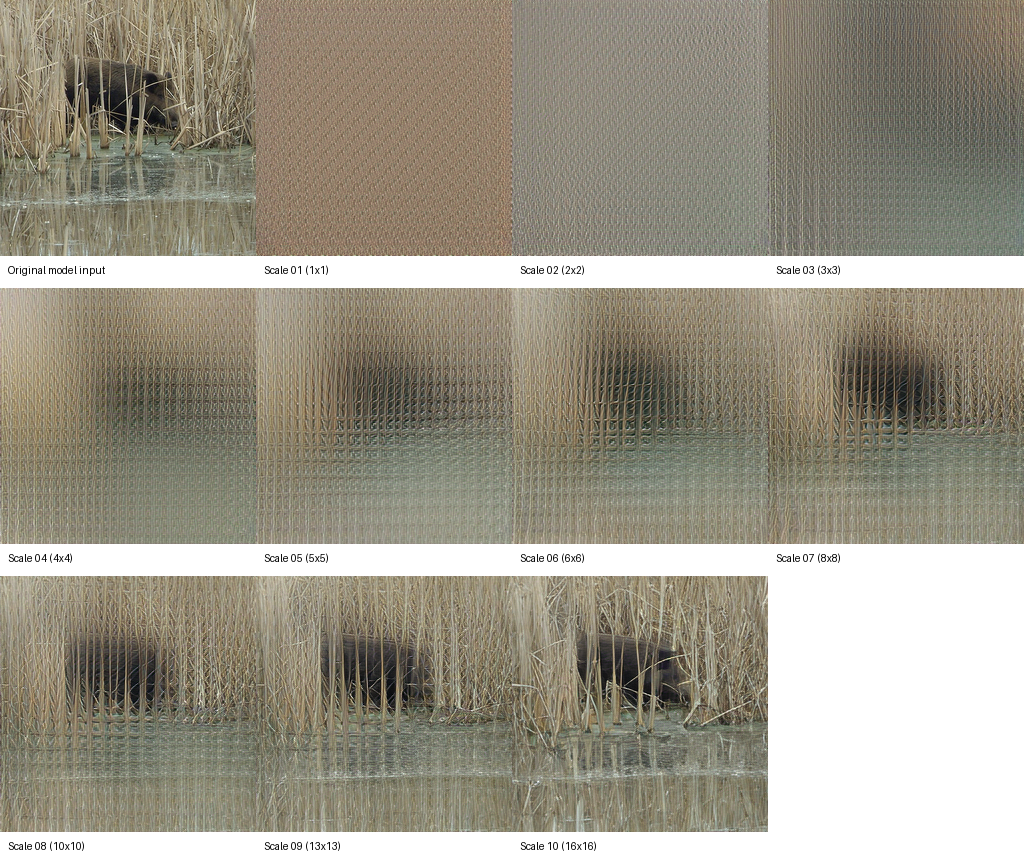}
    \caption{Coarse-to-fine reconstruction trajectory of a real image from ImageNet.}
    \label{fig:traj_real_imagenet}
\end{figure}

\begin{figure}[htbp]
    \centering
    \includegraphics[width=0.8\linewidth]{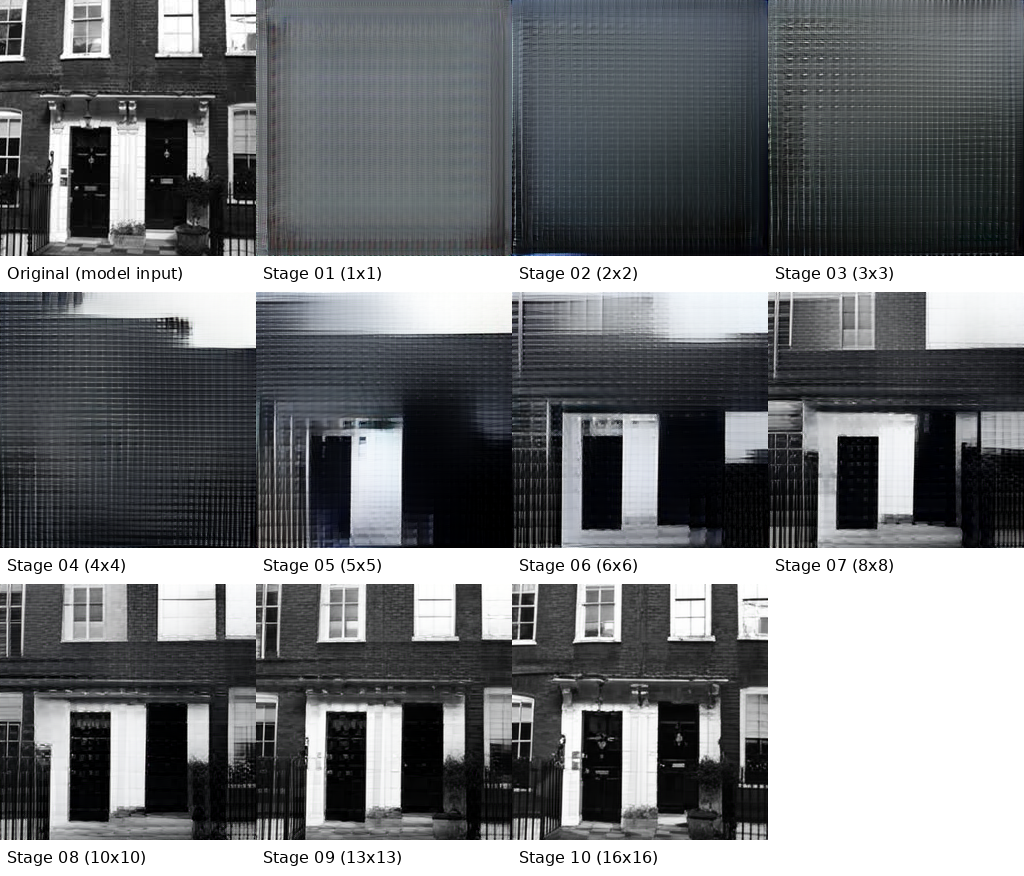}
    \caption{Coarse-to-fine reconstruction trajectory of a real image from LSUN.}
    \label{fig:traj_real_lsun}
\end{figure}

\begin{figure}[htbp]
    \centering
    \includegraphics[width=0.8\linewidth]{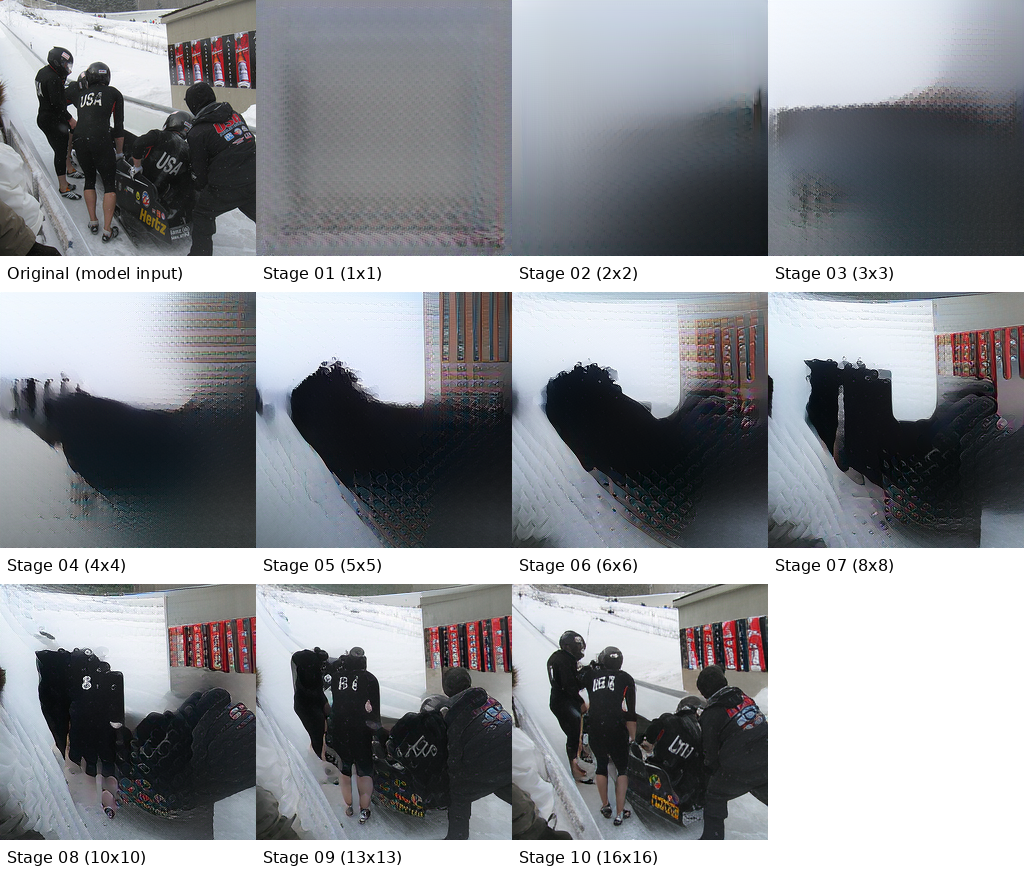}
    \caption{Coarse-to-fine reconstruction trajectory of a real image from MS COCO.}
    \label{fig:traj_real_coco}
\end{figure}

\begin{figure}[htbp]
    \centering
    \includegraphics[width=0.8\linewidth]{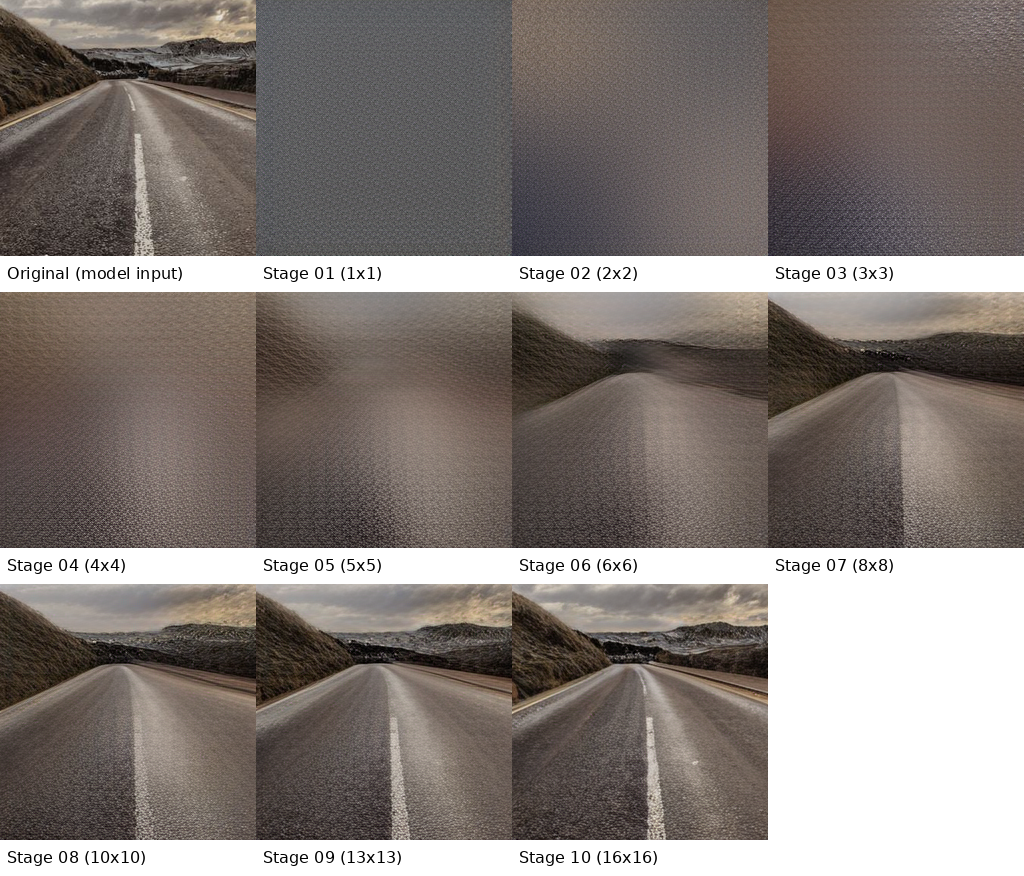}
    \caption{Coarse-to-fine reconstruction trajectory of a real image from Chameleon.}
    \label{fig:traj_real_chameleon}
\end{figure}

\begin{figure}[htbp]
    \centering
    \includegraphics[width=0.8\linewidth]{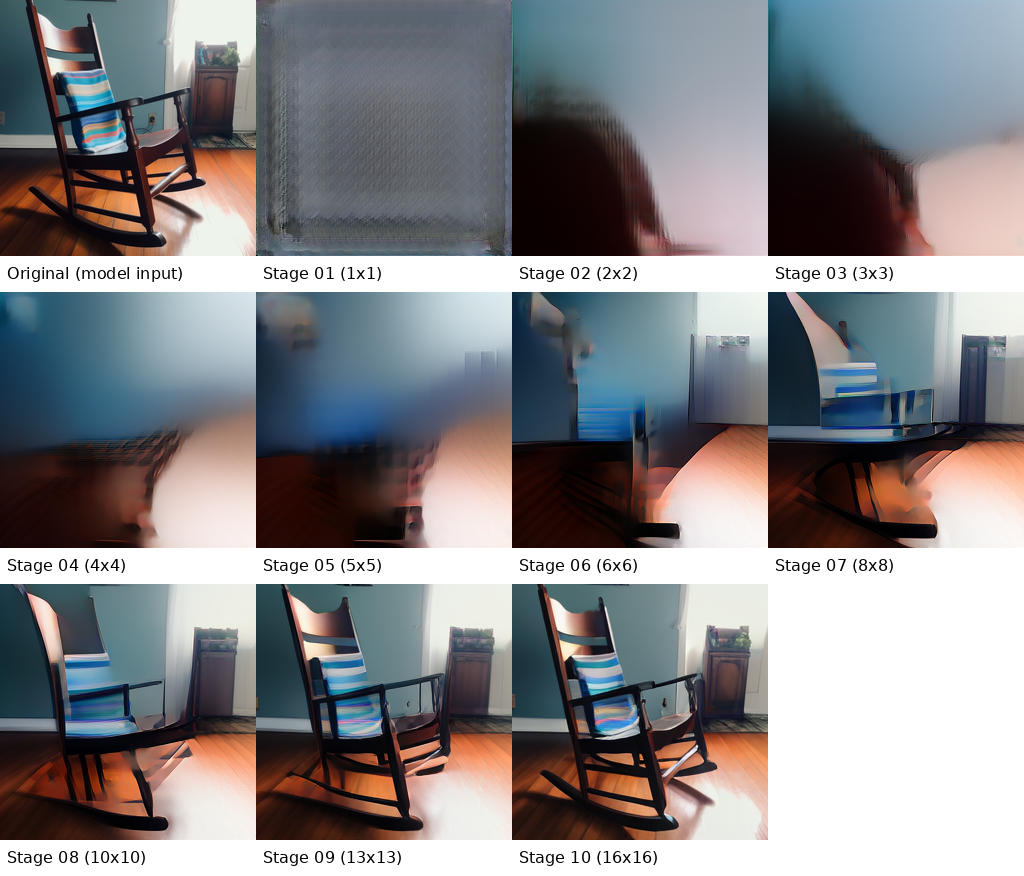}
    \caption{Coarse-to-fine reconstruction trajectory of an AI-generated image from GenImage.}
    \label{fig:traj_generated_genimage}
\end{figure}

\begin{figure}[htbp]
    \centering
    \includegraphics[width=0.8\linewidth]{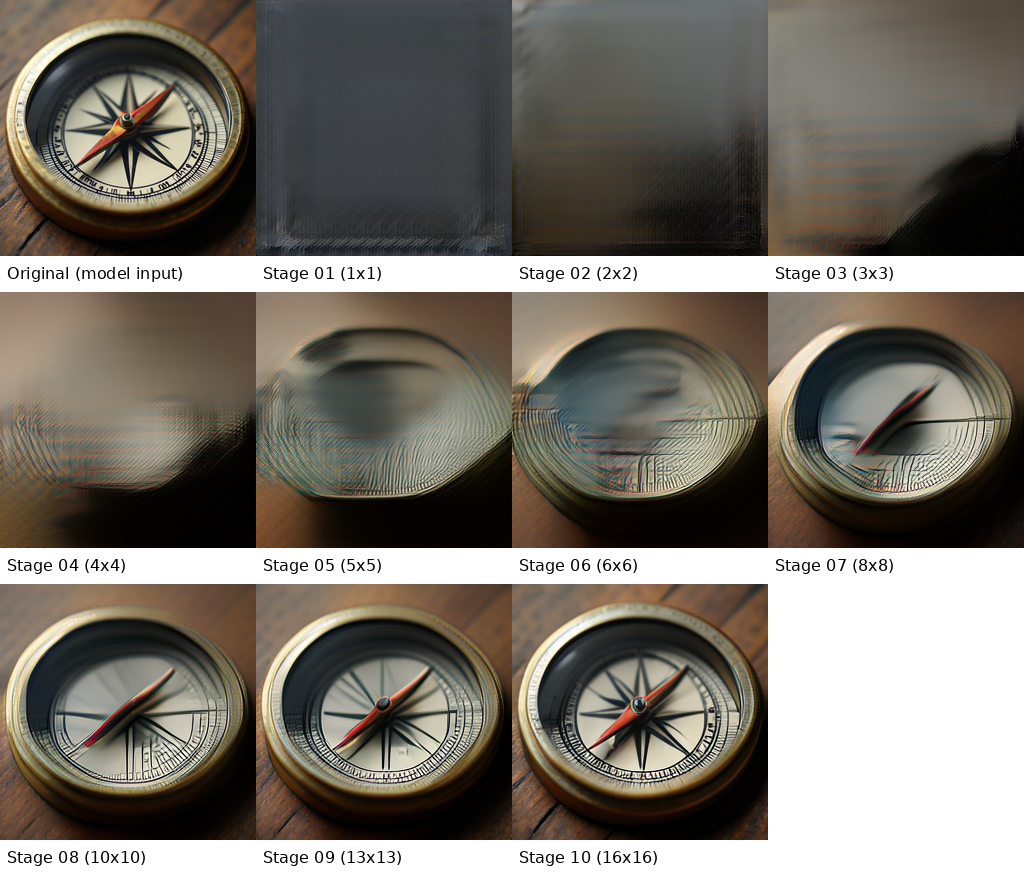}
    \caption{Coarse-to-fine reconstruction trajectory of an AI-generated image from ARForensics.}
    \label{fig:traj_generated_arforensics}
\end{figure}

\begin{figure}[htbp]
    \centering
    \includegraphics[width=0.8\linewidth]{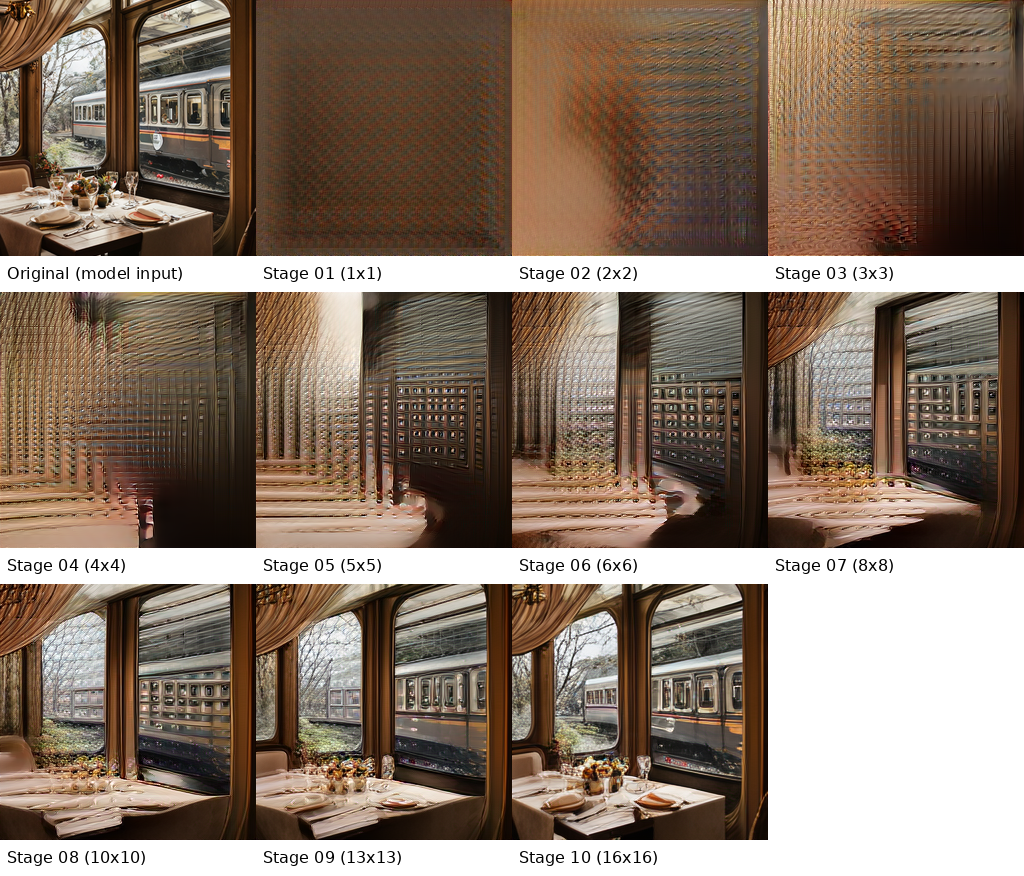}
    \caption{Coarse-to-fine reconstruction trajectory of an AI-generated image from EvalGEN.}
    \label{fig:traj_generated_evalgen}
\end{figure}

\begin{figure}[htbp]
    \centering
    \includegraphics[width=0.8\linewidth]{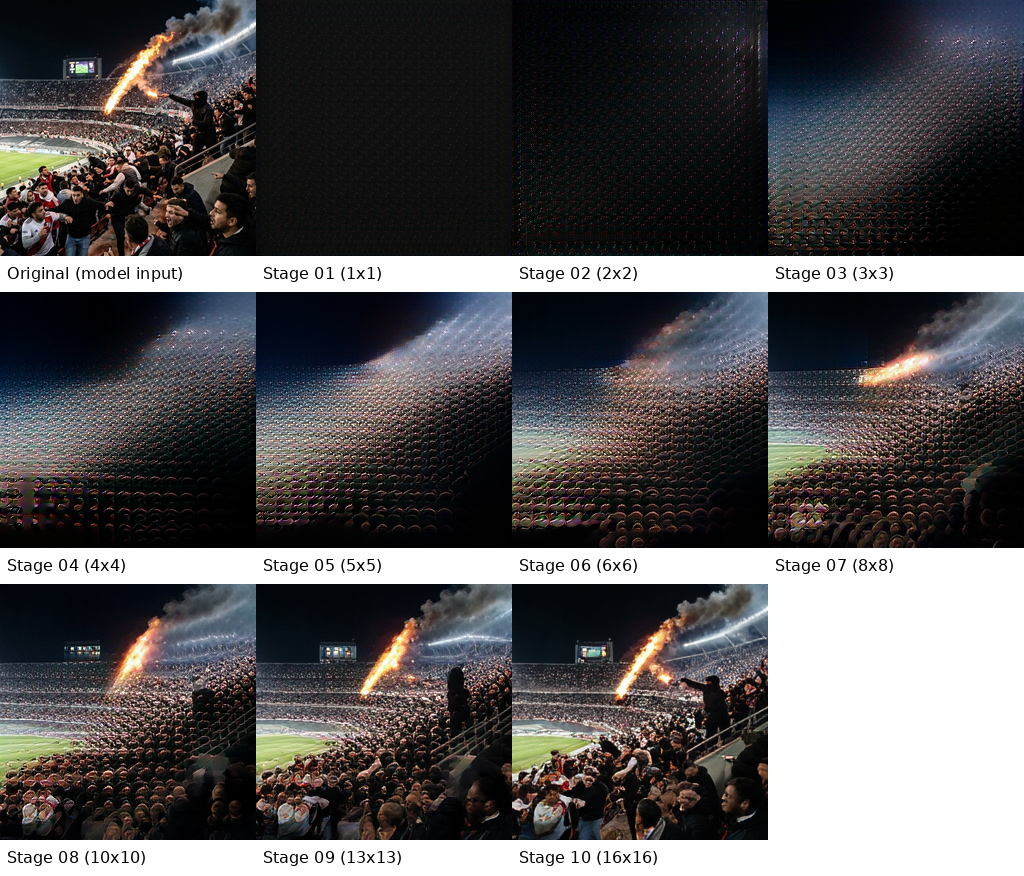}
    \caption{Coarse-to-fine reconstruction trajectory of an AI-generated image from Qwen-Image-Bench.}
    \label{fig:traj_generated_qwen}
\end{figure}

\begin{figure}[htbp]
    \centering
    \includegraphics[width=0.8\linewidth]{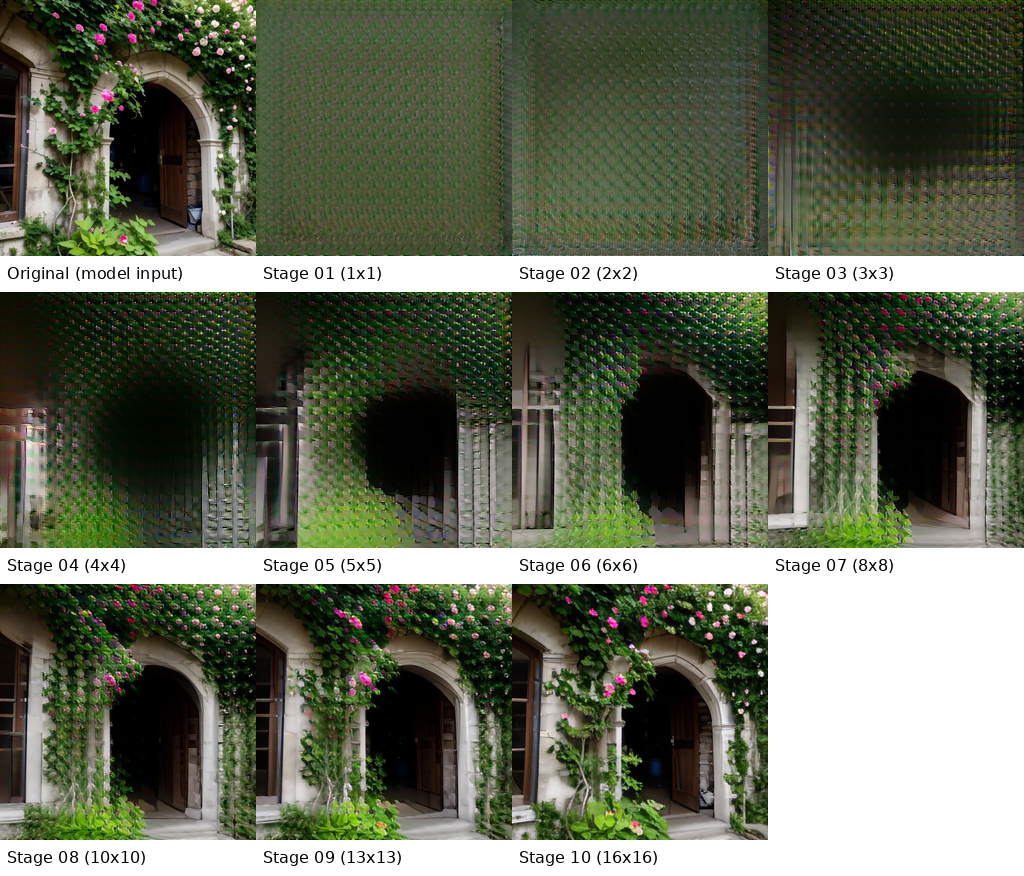}
    \caption{Coarse-to-fine reconstruction trajectory of an AI-generated image from Chameleon.}
    \label{fig:traj_generated_chameleon}
\end{figure}

\begin{figure}[htbp]
    \centering
    \includegraphics[width=0.8\linewidth]{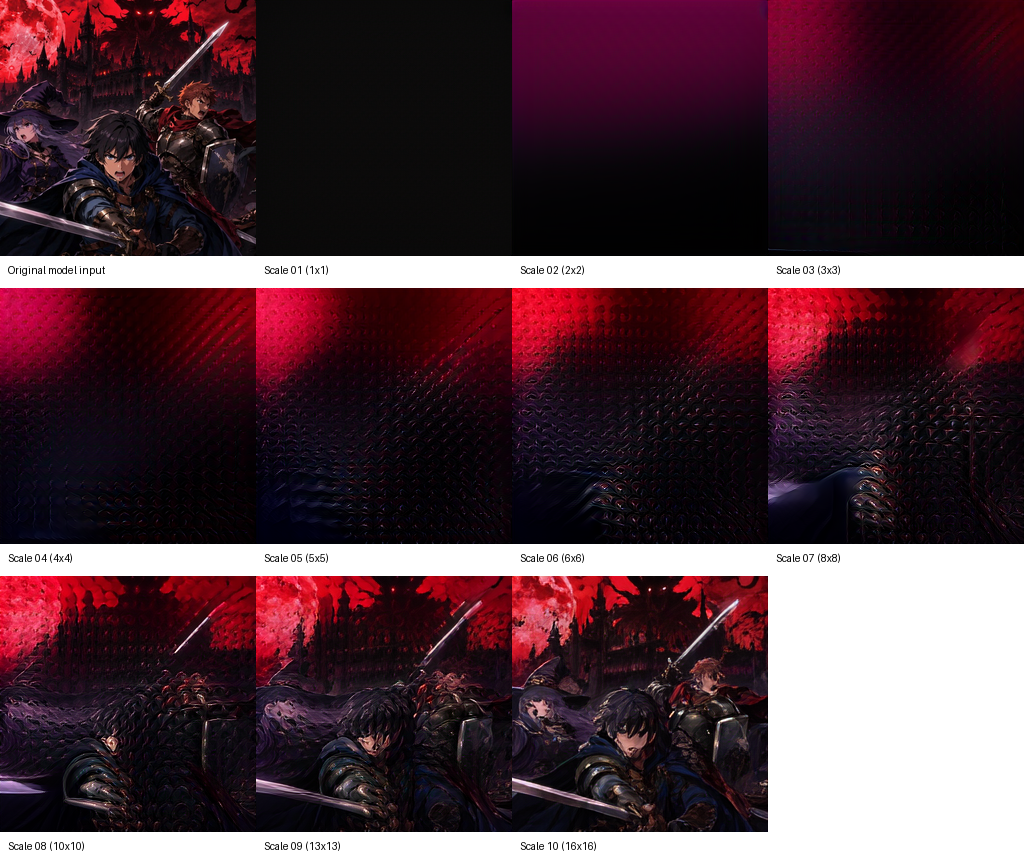}
    \caption{Coarse-to-fine reconstruction trajectory of an AI-generated image from GPT-Image-2-Twitter.}
    \label{fig:traj_generated_twitter}
\end{figure}

\end{document}